\documentclass{article}

\usepackage{microtype}
\usepackage{graphicx}
\usepackage{subcaption}

\usepackage{booktabs} 

\usepackage{hyperref}

\usepackage[accepted]{icml2025}

\usepackage{amsmath}
\usepackage{amssymb}
\usepackage{mathtools}
\usepackage{amsthm}

\usepackage[capitalize,noabbrev]{cleveref}

\usepackage{multirow,multicol,makecell}

\theoremstyle{plain}
\newtheorem{theorem}{Theorem}[section]

\theoremstyle{definition}

\theoremstyle{remark}
\newtheorem{remark}[theorem]{Remark}

\usepackage[textsize=tiny]{todonotes}

\icmltitlerunning{Broadband Ground Motion Synthesis by Diffusion Model with Minimal Condition}

\begin{document}

\twocolumn[
\icmltitle{Broadband Ground Motion Synthesis by Diffusion Model \\with Minimal Condition}

\icmlsetsymbol{equal}{*}

\begin{icmlauthorlist}
\icmlauthor{Jaeheun Jung}{equal,aimlk}
\icmlauthor{Jaehyuk Lee}{equal,aimlk}
\icmlauthor{Changhae Jung}{aimlk}
\icmlauthor{Hanyoung Kim}{aimlk}
\icmlauthor{Bosung Jung}{aimlk}
\icmlauthor{Donghun Lee}{aimlk}

\end{icmlauthorlist}

\icmlaffiliation{aimlk}{Department of Mathematics, Korea University, 145 Anam-ro, Seongbuk-gu, Seoul, Republic of Korea}

\icmlcorrespondingauthor{Donghun Lee}{holy@korea.ac.kr}

\icmlkeywords{Machine Learning, ICML}

\vskip 0.3in
]

\printAffiliationsAndNotice{\icmlEqualContribution} 

\begin{abstract}

Shock waves caused by earthquakes can be devastating. 
Generating realistic earthquake-caused ground motion waveforms help reducing losses in lives and properties, yet generative models for the task tend to generate subpar waveforms. 
We present High-fidelity Earthquake Groundmotion Generation System (HEGGS) and demonstrate its superior performance using earthquakes from North American, East Asian, and European regions. 
HEGGS exploits the intrinsic characteristics of earthquake dataset and learns the waveforms using an end-to-end differentiable generator containing conditional latent diffusion model and hi-fidelity waveform construction model. 
We show the learning efficiency of HEGGS by training it on a single GPU machine and validate its performance using earthquake databases from North America, East Asia, and Europe, using diverse criteria from waveform generation tasks and seismology. 
Once trained, HEGGS can generate three dimensional E-N-Z seismic waveforms with accurate P/S phase arrivals, envelope correlation, signal-to-noise ratio, GMPE analysis, frequency content analysis, and section plot analysis.

\end{abstract}

\section{Introduction}

Broadband ground motion caused by seismic waves is crucial in the study of earthquakes and geology since it includes important features related to subsurface structures of the solid Earth.
At the same time, it is a great challenge from signal processing perspective, as observed ground motion waveform signals cover a wide frequency band and are caused by rare and unevenly distributed earthquake events. 

As the size of systematically recorded seismic waveforms grew, various improvements in seismological applications were made by analyzing historically observed seismic waveforms. 
For example, the accuracy of earthquake analysis was improved, early warning systems for earthquake-prone areas were polished, and earthquake-resistant architectural designs became more robust.
Recently deep learning found successful applications in seismology \cite{DLseismology}, 
such as seismic signal denoising \cite{8802278}, fault recognition \cite{AN2021104776}, and earthquake event detection \cite{mousavi2020earthquake,saad2023eqcct}. 

However, the field still faces a shortage of data, particularly for large-scale earthquakes as they are rare \cite{GANO, katsanos2010selection}.
Recently, deep-learning based synthesis of seismic waveforms has emerged as a potential solution, mostly employing GAN-based generative models conditioned with various geological and seismological information \cite{wang2021seismogen,BBGAN,conseisgen,phasegen}.
However, the synthesized waveforms from these models often lack seismological realism, such as phase arrival times and amplitude of ground motions.

We see this problem as a mixed artifact of conditioned generation model architecture and unreliable conditioning information. 
Hence, we propose adapting diffusion model architecture \cite{diffusion15,DDPM}, which have shown superior realism and stability in image generation, to seismic waveform data in order to present a novel generation system for seismologically realistic ground motions with bare minimum set of conditioning information.

\subsection*{Our Contribution}
\begin{itemize}
    \item We design a novel seismic waveform generation model, HEGGS, that requires a bare-minimum set of conditional information on the earthquake and the observation point.
    \item We demonstrate constructing datasets for HEGGS using openly available seismic datasets, such that observed waveforms are paired with the source earthquake and time-aligned to the earthquake origin time.
    \item We design an end-to-end training method for the model and demonstrate its effectiveness by learning to generate high-fidelity earthquake ground motions from earthquakes in North America, East Asia, and Europe. 
    \item We validate the superior fidelity of generated samples from HEGGS against benchmark models in various perspectives including seismology-inspired metrics such as GMPE analysis and phase arrival prediction.
\end{itemize}

\section{Key Idea}
\label{sec:prob_state}

\begin{figure*}[th]
  \centering 
  \includegraphics[width=\textwidth]{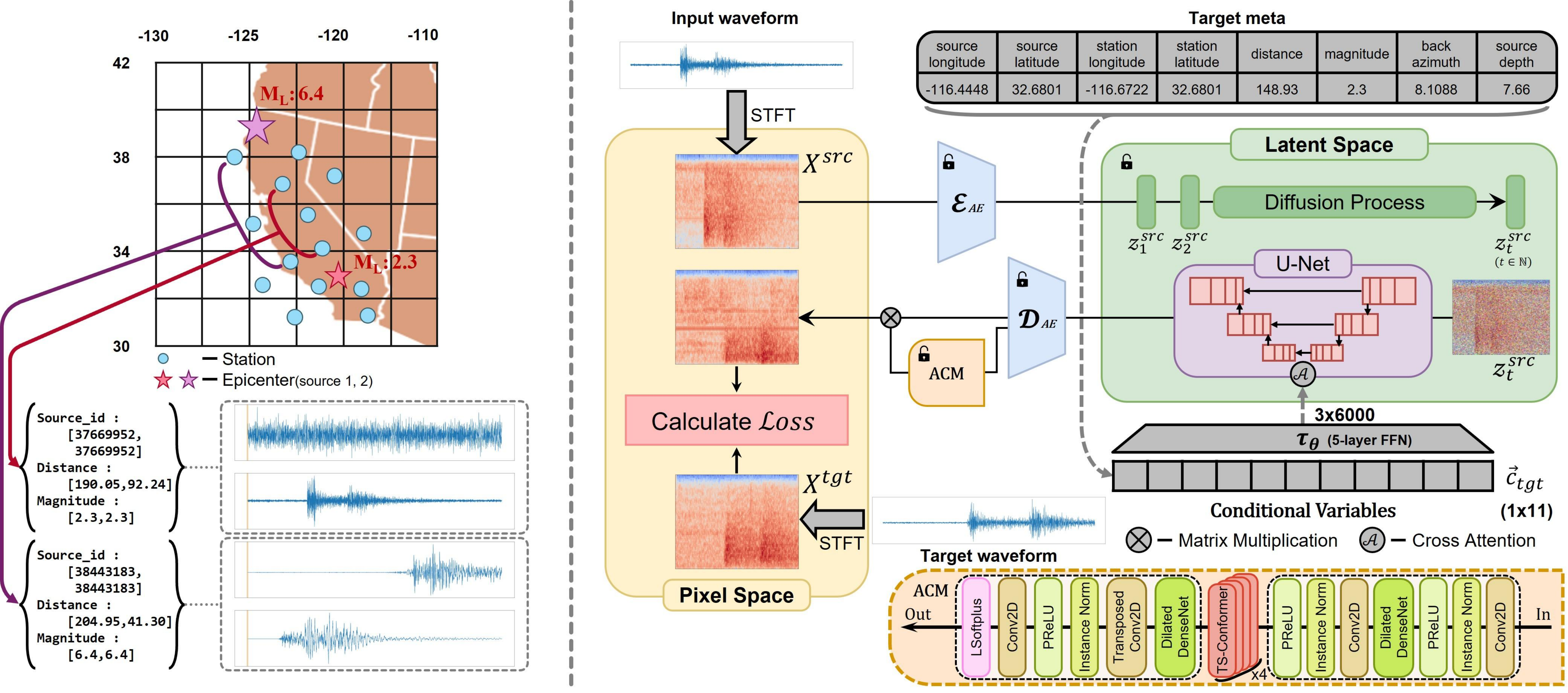}
  \caption{Left : Visualization of SCEDC data using paired waveforms. It shows that earthquake events can be detected at greater distances depending on magnitude. Right : Diagram of the waveform generative model architecture of HEGGS and its training loss.
  }\label{fig:Data_and_Architect}
\end{figure*}

Our goal is to robustly synthesize the broadband ground motion data with high level of seismological realism when compared to actual seismograph-recorded waveforms caused by earthquake events.
We cast this seismic waveform synthesis problem into conditional waveform generation task to learn from seismic waveform databases, with minimal dependency on conditional information. 

We use the following as the minimal conditional info:
\begin{enumerate}
    \item $s_{lat}, s_{lon}$ : latitude and longitude of the station to observe the waveform data. 
    \item $e_{lat}, e_{lon}$ : latitude and longitude of the hypocenter.
    \item $e_{dep}$ : depth of the hypocenter, in kilometers.
    \item $M_L$ : local magnitude of the earthquake.
\end{enumerate}
This set of information is usually considered insufficient for ground motion synthesis. 
For example, seismological properties of the source earthquake such as focal mechanism or local geological properties of observation point such as $V_{S30}$ are often required in prior works. 
Instead of additionally demanding the often expensive-to-obtain info, we desire a generative model that learns directly from seismic waveforms with \emph{minimal conditional info} as metadata.

When an earthquake happens, its shock waves propagate and are recorded by nearby seismographs as three-dimensional seismic waveforms named seismograms. 
Naturally, these seismograms are correlated by the common information about the source earthquake as well as the information specific to each of the seismographs such as geological conditions around the observation location.
Regional seismic waveform datasets contain multiple waveform observations that can be paired to a source earthquake event, as illustrated in the left panel of \cref{fig:Data_and_Architect}.
The observations paired to the same earthquake would share properties of the same source earthquake, \emph{while containing different information pertaining to their respective observation location}. 
This is how we liberate ourselves from asking additional conditional info.

We exploit this intrinsic pair-ability of the seismic waveform datasets, and construct paired waveform-metadata datasets from three earthquake databases from different continents: INSTANCE \cite{instance} from Europe, KMA \cite{h_magnitude} from East Asia, and SCEDC \cite{scedc} from North America.
Raw waveforms and corresponding metadata including locations, earthquake ID, magnitude and earthquake occurrence time are collected, 
and the processing of each datasets are detailed in \cref{sec:appendix-dataset}.

\section{Method} 
\label{sec: method}

Inspired by a conditional music generation method \cite{tango}, our method first creates spectrograms with a diffusion model and then convert spectrograms into waveforms. 
Our generative model fully exploits the pair-ability of seismic waveform datasets shown in \cref{sec:prob_state} to train both the diffusion process for spectrogram generation and the high-fidelity decoder for waveform generation. 
We name the method HEGGS, an acronym for High-fidelity Earthquake Groundmotion Generation System.

\subsection{Pair-Exploiting Diffusion Model}

For each earthquake event, we sample a pair of waveforms $(W^{src},W^{tgt})$ from dataset and convert it to spectrograms $(X^{src}, X^{tgt})$ and construct conditional vector of target station $\Vec{c}_{tgt}$ by preprocessing. 

Let $q(x_{1:T};X)$ be the forward process of the diffusion model, and consider two trajectories $q(x_t^{src}|X^{src})$ and $q(x_t^{tgt}|X^{tgt})$. Recall that $X^{src}$ and $X^{tgt}$ shares the property of earthquake, we may assume that from $X^{src}$ and $\Vec{c}_{tgt}$ we can gather enough features of earthquake to generate $X^{tgt}$. In this approach, we may consider the transform map $\eta(x_t^{src},\Vec{c}_{tgt},t)$ for $t>0$ which maps the latent of input $X^{src}$ to the latent of target $X^{tgt}$ as a random variable, with following assumption:
\begin{equation}\label{eqn:src-to-tgt-eta}
    \eta(x_t^{src},\Vec{c}_{tgt},t) \sim q(x_t^{tgt}|X^{tgt}).
\end{equation}

Referring  \cite{vpred}, the loss function $\mathcal{L}_{DM}$ of diffusion model in $\mathbf{x}$-space (sample space) is:
\begin{equation}\label{eqn:loss_DM_vanilla}
    \mathcal{L}_{DM}=\mathbb{E}_{(X^{tgt},\Vec{c}_{tgt}),\epsilon,t} \|X^{tgt}-\mathbf{x}_\theta(x_t^{tgt},\Vec{c}_{tgt},t)\|^2.
\end{equation}
while the SNR weight is simplified.

Considering the \cref{eqn:src-to-tgt-eta}, we rewrite the loss function as 
\begin{equation}\label{eqn:loss_DM_paired}
    \mathcal{L}_{DM}' = \mathbb{E}_{(X^{src},X^{tgt},\Vec{c}_{tgt}),\epsilon,t}\|X^{tgt}-\mathbf{m}_\theta(x_t^{src},\Vec{c}_{tgt},t)\|^2
\end{equation}
where
\begin{equation}
    \mathbf{m}_\theta(x,\Vec{c},t) = \mathbf{x}_\theta(\eta(x,\Vec{c},t),\Vec{c},t).
\end{equation}
Hence, we predict $\mathbf{m}_\theta$ by neural network, which is a composition of latent transform function and denoising model.

For the sampling of the reverse process, we exploit the same procedure of the denoising process of diffusion, as 
\begin{equation}
    x_{t-1}^{tgt} = \tilde{\mu}_t(x_t^{tgt},\textbf{m}_\theta(x_t^{tgt},\vec{c}_{tgt},t))+\sigma_t\textbf{z}, \textbf{z}\sim N(0,I)
\end{equation}
where $\tilde{\mu}_t(x_t,x_0)$ is mean vector of $q(x_{t-1}|x_t,x_0)$, introduced in Eq. (7) of \cite{DDPM}.

This is equivalent to conventional denoising process, as
\begin{equation}
    \eta(x_t^{tgt},\Vec{c}_{tgt},t) \,{\buildrel d \over =}\, x_t^{tgt}
\end{equation}
by assumption and thus
\begin{equation}
    \mathbf{m}_\theta(x_t^{tgt},\Vec{c}_{tgt},t)=\mathbf{x}_\theta(x_t^{tgt},\Vec{c}_{tgt},t).
\end{equation}

Therefore, pair-exploiting training process of HEGGS allows the diffusion model to generate $X^{tgt}$ from the Gaussian noise $x_T^{tgt}\sim \mathcal{N}(0,I)$ following conventional reverse process with $\mathbf{m}_\theta$.

\subsection{End-to-end Model Training}

From the idea of LDM  \cite{LDM}, we consider the autoencoder comprised of a downsampling module $\mathcal{E}_{AE}$ and an upsampling module $\mathcal{D}_{AE}$, and construct diffusion model on latent space with smaller dimension. 
If there were a suitable pretrained autoencoder, the LDM loss would be
\begin{equation}\label{eqn:loss-ldm}
    \mathcal{L}_{LDM}' = \mathbb{E}_{(Z^{src},Z^{tgt},\vec{c}_{tgt}),\epsilon,t} \|Z^{tgt}-\mathbf{m}_\theta(z_t^{src},\vec{c}_{tgt},t)\|^2
\end{equation}
where $Z=\mathcal{E}_{AE}(X)$, $z_t^{src}$ is latent of diffusion process of $Z^{src}$.

There is no suitable encoder-decoder model for seismic waveforms, so we modify \cref{eqn:loss-ldm} into an end-to-end loss function as shown below:
\begin{equation}\label{eqn:loss-ours}
    \begin{aligned}
        &\mathcal{L}_{ours}  \\&:=\mathbb{E}_{(X^{src},X^{tgt},\vec{c}_{tgt}),\epsilon,t} \|X^{tgt}-\mathcal{D}_{AE}(\mathbf{m}_\theta(z_t^{src},\vec{c}_{tgt},t))\|^2
    \end{aligned}
\end{equation}
where $z_t^{src}=\sqrt{\overline{\alpha}_t}\mathcal{E}_{AE}(X^{src})+\sqrt{1-\overline{\alpha}_t}\epsilon$ .

Using $\mathcal{L}_{ours}$ as the loss function, we train the waveform generation model end-to-end, covering the encoder, the diffusion module, and the decoder with ACM (Amplitude Correction Module), as shown in the right panel of \cref{fig:Data_and_Architect}.
For the detailed implementation in the diffusion module, we used a U-Net backbone for $\mathbf{m}_\theta$, brought $\mathcal{E}_{AE}$ and $\mathcal{D}_{AE}$. 
More details on the specifications of HEGGS and its training recipe can be found in \cref{sec:appendix-implementation}. 

After training diffusion model with HEGGS, we generate waveform with conventional reverse process by setting $z_T^{tgt}$ by Gaussian noise or $Z^{src}$. The details with pseudocode of training and generation, can be found in \cref{appendix:pair-exploiting}.

\section{Empirical Verification}

We showcase the performance of HEGGS in three cases: 
\begin{enumerate}
    \item generate waveforms of existing earthquakes at existing observation stations $\vec{c}_{tgt}$, using an observed earthquake information $W^{src}$ as input waveform.
    \item generate at arbitrary locations $\vec{c}_{tgt}'$,  using an observed earthquake information $W^{src}$ as input waveform. 
    \item generate waveforms of fictitious earthquake information $\vec{c}_{tgt}^{''}$ (also, without $W^{src}$).
\end{enumerate}

The first case is designed to verify the fidelity of HEGGS, by comparing the generated samples to ground truth waveforms. 
We present quantitative results in \cref{sebsec:quantitative_eval} and qualitative analyses with visualizations in \cref{subsec:qualitative_waveform,subsec:qualitative_spectrogram,subsec:qualitative_freqcontest}. 
The results from the second case are presented throughout   \cref{sebsec:quantitative_eval,sec:qualitative}, and in \cref{subsec:qualitative_magnitude} we show the results from the third case.

\subsection{Quantitative Evaluation}
\label{sebsec:quantitative_eval}
To assess and compare the effectiveness of models synthesizing seismic waves, we conducted a comprehensive quantitative analysis focusing on key parameters including P-wave and S-wave arrival times, 
GMPE analysis
, and similarity measures such as envelope correlation, spectrogram image similarity, as well as signal-to-noise ratio (SNR), and peak signal-to-noise ratio (PSNR).
Specifically, we compare generated synthetic waveform $W^{pred}$ from $\Vec{c}_{tgt}$ and compare it with corresponding ground truth waveform $W^{tgt}$ to compute each metric. 

For comparison, we also trained the following benchmark models on SCEDC: Seismogen \cite{wang2021seismogen}, Conseisgen \cite{conseisgen}, BBGAN \cite{BBGAN} and LDM \cite{LDM}. Since the input shape of waveform $W^{tgt}$ or spectrogram $X^{tgt}$ is different from each of the models, we give reasonable modifications to them for the training and evaluation. The detailed changelogs are listed in \cref{sec:appendix-benchmark}.

\subsubsection{Phase Arrival Times}\label{subsec:result-PSarrival}

The arrival times of P-wave and S-wave are the most basic, but important properties of seismic waveforms, as determining P-wave and S-wave from the seismogram is the first step of earthquake analysis. 
Since we want to assess the fidelity of the generated waveforms, we fine-tune the SeisBench \cite{seisbench} implementation of EQTransformer (EQT)  \cite{mousavi2020earthquake} on each dataset to use it in labeling phase arrivals times and compare them with the ground truth arrival times. 
The detailed training recipe we used for EQT is given in \cref{sec:appendix-eqt}. 
We present the performance measure of the finetuned EQT in the test dataset in \cref{tab:EQT}. Note that the waveforms are inherently normalized as preprocessing step of EQT prediction.
\begin{table}[th] 
    \caption{Performance of EQT picker on each test dataset. Samples with errors less than 0.5s are considered to be positive for F1 score computation.}\label{tab:EQT}
    \vspace{0.26cm}
  \centering
  \begin{tabular}{c@{\hskip3pt}cccc@{\hskip4pt}c}
    \toprule
    Dataset    & P\_MAE(s)&S\_MAE(s) & P\_F1 & S\_F1 \\
    \midrule
    SCEDC& 0.1116&0.2189&0.9728&0.9384\\
    KMA&0.0993&0.1362&0.9635&0.9624\\
    INSTANCE&0.1738&0.3151&0.9797&0.9099\\
    \bottomrule
  \end{tabular}
\end{table}

\begin{table*}[t]  
  \centering
  \caption{Results of quantitative analysis. Models were evaluated with $W^{src}$ when it is trained with paired data, otherwise without $W^{src}$, except (*): evaluated without $W^{src}$, while the model was trained with paired data. }\label{tab:quantitative_result}
  \vspace{0.26cm}
  \begin{tabular}{c@{\hskip3pt}c@{\hskip3pt}c@{\hskip3pt}c@{\hskip3pt}c@{\hskip3pt}c@{\hskip3pt}c@{\hskip6pt}c@{\hskip3pt}c}
    \toprule
        \multirow{2}{*}{Dataset}&\multirow{2}{*}{Model}&\multirow{2}{*}{Input}&\multicolumn{5}{c}{Waveform} &\multicolumn{1}{c}{Spectrogram}\\
    \cmidrule(r){4-8}\cmidrule(r){9-9}
 &&& P\_MAE (s) & S\_MAE (s) & $env.corr$ & SNR & PSNR & MSE \\
    \midrule
    \multirow{10}{*}{SCEDC}&\multirowcell{2}{SeismoGen \\ \cite{wang2021seismogen}}&w/o $W^{src}$ & 1.9558 & 3.6246 & 0.4895 & -8.6166 & 23.5431 & 1.4124 \\
    &&w/  $W^{src}$& 1.8426 & 3.3325& 0.5454 & -8.6282 & 23.6354 & 0.8063 \\
    \cmidrule(r){3-9}
    &\multirowcell{2}{ConSeisGen \\ \cite{conseisgen}}&w/o $W^{src}$ & 3.9724 & 6.8992 & 0.3246 & -8.6216 & 23.6416 & 0.7461 \\
    &&w/  $W^{src}$& 3.9102 & 6.8055 & 0.2980 & -8.5341 & 23.5329 & 0.9356\\
    \cmidrule(r){3-9}
    &\multirowcell{2}{BBGAN \\ \cite{BBGAN}} &\multirow{1}{*}{w/o $W^{src}$}& \multirow{1}{*}{6.4210}  & \multirow{1}{*}{10.416}  & \multirow{1}{*}{0.1950} & \multirow{1}{*}{-3.0093} & \multirow{1}{*}{23.7598}  & \multirow{1}{*}{1.6150} \\
    &&w/  $W^{src}$&\multicolumn{6}{c}{diverged}\\
    \cmidrule(r){3-9}
    &\multirowcell{2}{LDM \\ \cite{LDM}}&w/o $W^{src}$ & 1.1142  & 1.7294  & 0.6932 & -3.0202 & \textbf{24.7573} & 0.2838 \\
    &&w/  $W^{src}$ & 0.5633  & 0.7808  & 0.7726 & -3.0015 & 19.6269  & 0.2426 \\
    \cmidrule(r){2-9}
    &\multirowcell{2}{HEGGS \\ (ours)} & (*)w/o $W^{src}$ & 0.5025 & 0.8003 & 0.7963 & -2.9891 & 24.6816  & 0.1531\\
    &&w/  $W^{src}$& \textbf{0.4760} & \textbf{0.5476} & \textbf{0.8187} &  \textbf{-2.0051} & 24.6553  & \textbf{0.1512} \\
    \midrule
    \multirow{4}{*}{KMA}&\multirowcell{2}{LDM \\ \cite{LDM}}&w/o $W^{src}$ & 1.6233  & 2.1125  & 0.7703 & -3.0006 & 25.3883 & 0.3521 \\
    &&w/  $W^{src}$ & 1.3521  & 1.6845  & 0.8076 & -2.9989 & 26.3658  & 0.3785 \\
    \cmidrule(r){2-9}
    &\multirowcell{2}{HEGGS \\ (ours)}&(*)w/o $W^{src}$&0.2988&0.5551&0.8769&-3.0034&26.2769&0.1867\\
    &&w/ $W^{src}$&\textbf{0.2763}&\textbf{0.4644}&\textbf{0.8785}&\textbf{-2.9768}&\textbf{26.8730}&\textbf{0.1682}\\
    \midrule
    \multirow{4}{*}{INSTANCE}&\multirowcell{2}{LDM \\ \cite{LDM}}&w/o $W^{src}$ & 0.8417  & 0.7847  & 0.7921 & -3.0062 & 22.0767 & 0.2927 \\
    &&w/  $W^{src}$ & 0.8187  & 0.7875  & 0.7898 & \textbf{-2.9904} & 22.0956  & 0.2841 \\
    \cmidrule(r){2-9}
    &\multirowcell{2}{HEGGS \\ (ours)}&(*)w/o $W^{src}$&0.5192&0.6804&0.8299&-3.0004&22.3690&0.1376\\
    &&w/ $W^{src}$&\textbf{0.5085}&\textbf{0.6378}&\textbf{0.8301}&-2.9976&\textbf{22.6870}&\textbf{0.1308}\\
    \bottomrule
  \end{tabular}
  \vspace{-0.3cm}
\end{table*}
Using the finetuned EQT as the phase picker, we label generated synthetic waveforms at a random station for every earthquake event from the test dataset.
The mean absolute error (MAE) metrics of the P wave and the S wave phase arrival times from the ground truth labels are reported 
in the first two columns of \cref{tab:quantitative_result}. 
The resulting phase arrival times of the P wave and the S wave are considered to be close to the ground-truth arrival times on all datasets, as the MAE values are measured to be significantly small while other benchmark models failed to generate earthquake event signals on correct arrival time. 
Notably, generating with input waveform $W^{src}$ gives better results compared to generating without $W^{src}$ (i.e. with noise), as the observation $W^{src}$ contains earthquake-specific information.

\subsubsection{Similarity Measures}\label{subsec:result-similarity}

We also compare the synthesized waveforms and corresponding spectrogram directly to the ground truth waveforms.
We use general-purpose similarity measures: the envelope correlation, SNR and PSNR for the waveforms, and MSE for the spectrograms.

Envelope correlation was calculated to measure the similarity between the envelopes of synthesized and observed seismic waves, providing insights into the overall waveform fidelity. We applied Savitzky-Golay Filtering \cite{savitzky1964smoothing} technique with polyorder $3$ before calculating the envelope correlation. In implementation, we exploit \cite{beyreuther2010obspy} implementation of cross correlation, which includes the waveform normalization.
Furthermore, SNR and PSNR metrics were employed to evaluate the quality of the synthesized seismic waves in terms of signal clarity and fidelity to the original data. 
Additionally, we compare the synthesized spectrogram $X^{syn}$ and spectrogram of observed seismic signals $X^{tgt}$ to quantify their similarity using image similarity. We normalized both spectrograms to compare 

The results are summarized in \cref{tab:quantitative_result}. The proposed method outperforms the benchmark models on almost all similarity metrics, which may imply that the generated $X^{pred}$ and $W^{pred}$ are more realistic in most cases. 
These quantitative analyses provide a comprehensive assessment of how similar HEGGS-generated waveforms are to the actual observed seismic ground motion.

\begin{figure*}[th]
\begin{subfigure}{0.3\linewidth}
  \centering
  \includegraphics[width=\columnwidth]{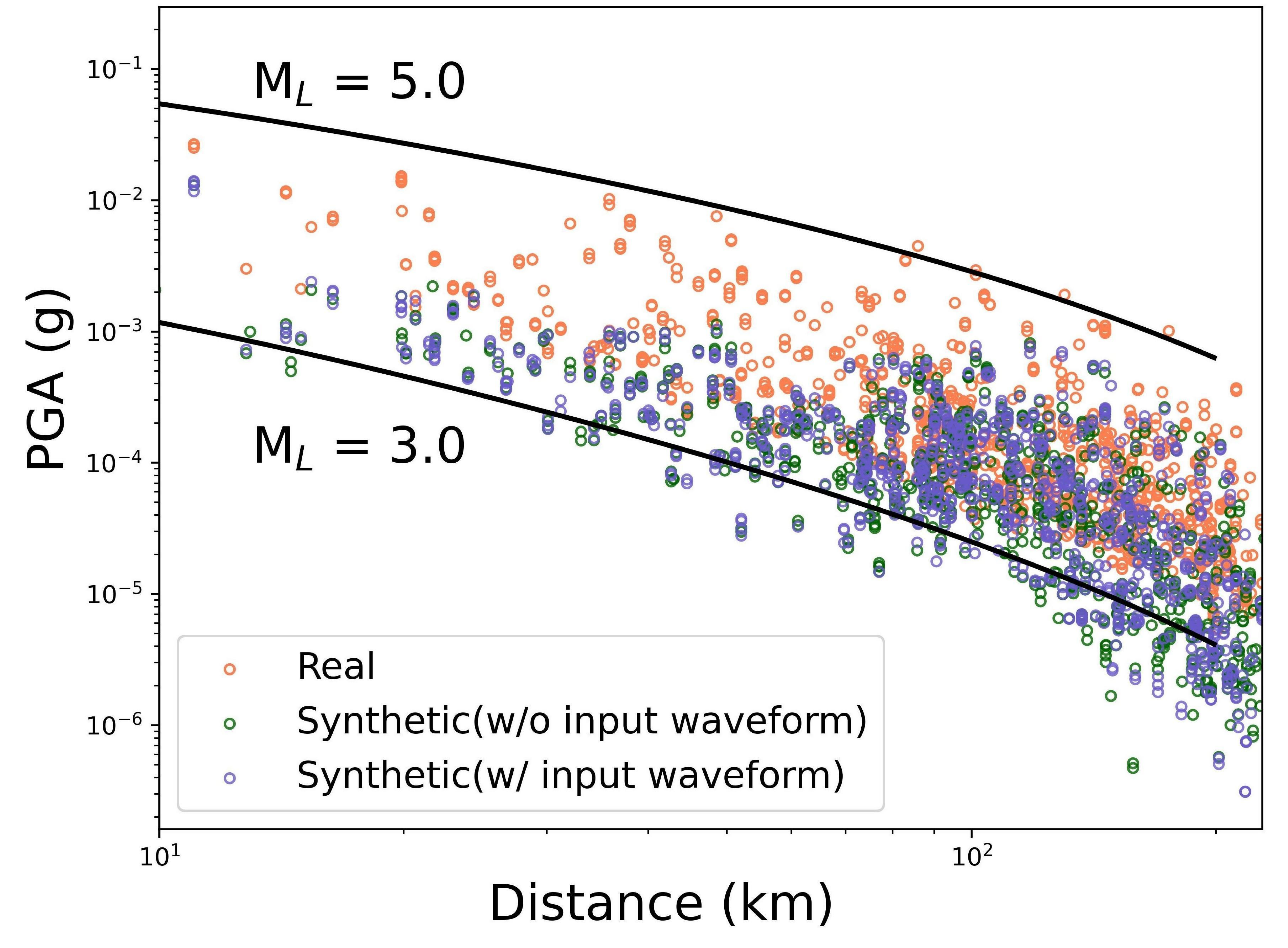}
  \vspace{-0.44cm}
  \caption{SCEDC}
  \label{fig:gmpe_scedc}
\end{subfigure}
\begin{subfigure}{0.3\linewidth}
  \centering
  \includegraphics[width=\columnwidth]{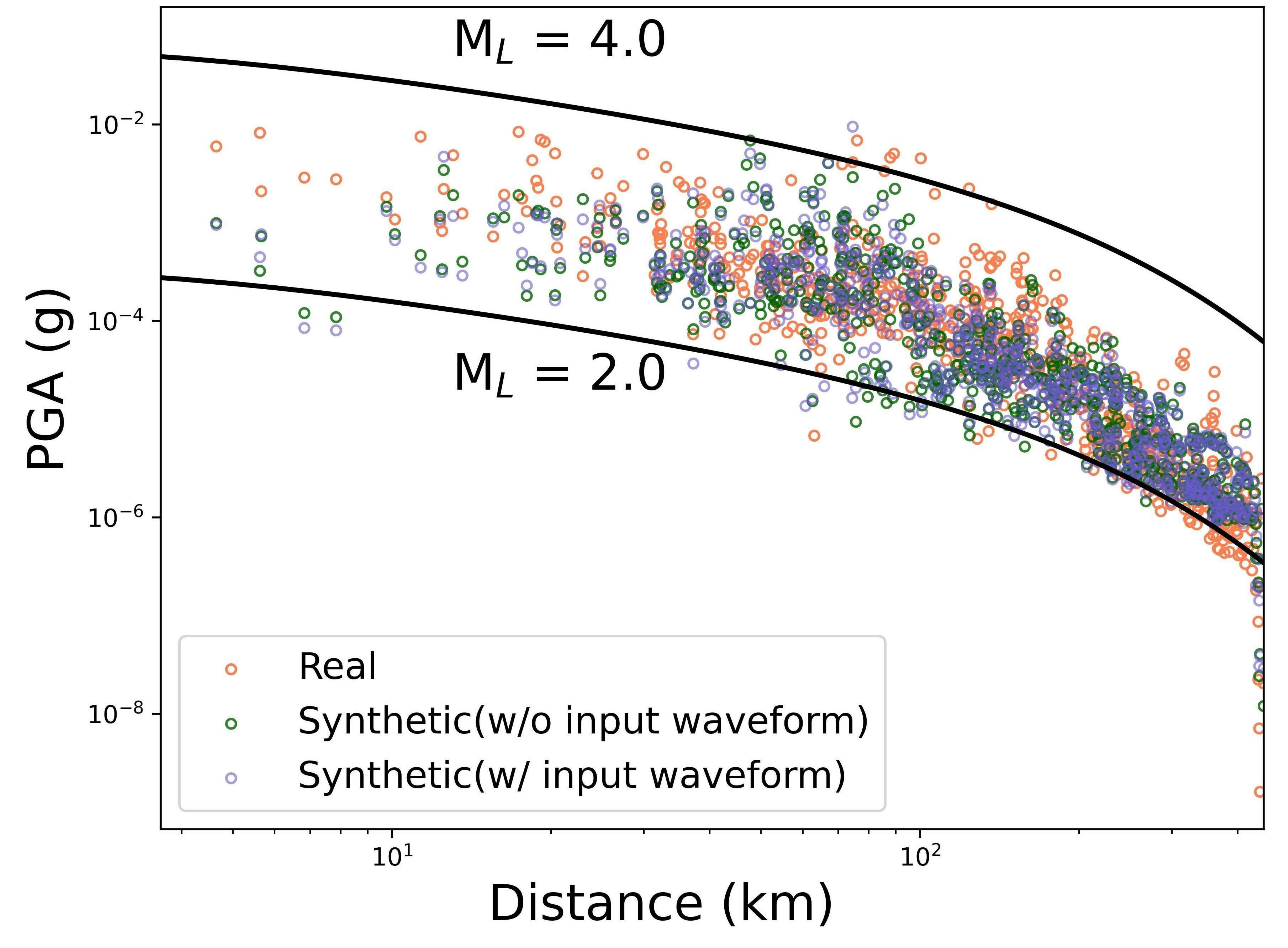}
  \vspace{-0.44cm}
  \caption{KMA}
  \label{fig:gmpe_kma}
\end{subfigure}
\begin{subfigure}{0.3\linewidth}
  \centering
  \includegraphics[width=\columnwidth]{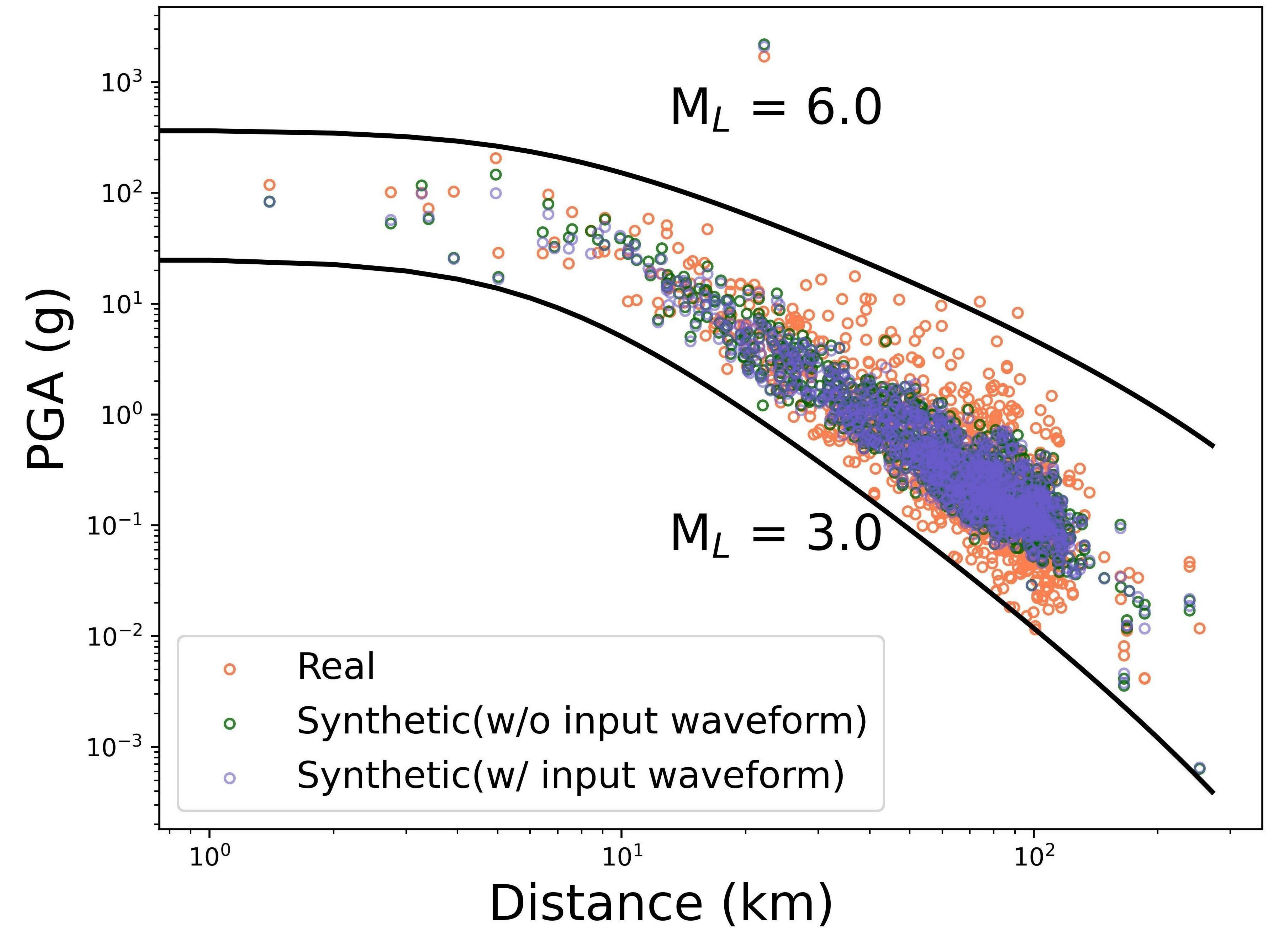}
  \vspace{-0.44cm}
  \caption{INSTANCE}
  \label{fig:gmpe_instance}
\end{subfigure}
  \centering
  \vspace{-0.26cm}
    \caption{
      Result of GMPE analysis in PGA values with respect to the distance. 
      The points in the figures represent the PGA values calculated from randomly selected waveforms from the test set containing earthquakes filtered by the magnitude range indicated by the black solid lines, and synthetic waveforms using the corresponding metadata.
      The subfigures correspond to the earthquake source: SCEDC (North America), KMA (East Asia), and INSTANCE (Europe).      
      }\label{fig:GMPE}
\end{figure*}
\subsubsection{GMPE Analysis}\label{subsec:result-GMPE}

Ground Motion Prediction Equation (GMPE) is a widely used mathematical modeling in seismology that predicts the ground shaking intensity caused by earthquakes, and it is crucial for seismic hazard assessment and earthquake-resistant structural engineering.
The GMPE model relates earthquake parameters, like local magnitude $M_L$ and hypocentral distance $R_{hypo}$, to ground motion characteristics, such as Peak Ground Acceleration (PGA). Since $M_L$ is obtained from the peak amplitude of the waveforms, the GMPE analysis shows how the scale of the generated waveforms from HEGGS matches the real observations.

Computing PGA  \cite{kr_gmpe,scedc_gmpe,instance_gmpe_a,instance_gmpe_b} is closely related to local magnitude $M_L$, which would be calculated by distinct formula \cite{h_magnitude,scedc_mag,inatance_mag} for each region. 
We follow the conventions in geoscience research to decide PGA computation formula, which is detailed in \cref{sec:appendix_gmpe}. 

GMPE analysis result, shown in \cref{fig:GMPE}, reveals that the synthetic seismic waveforms generated by HEGGS closely resemble how the observed ground truths in distribution. 
Also, the similarity in PGA values indicates that the magnitude of synthesized waveforms is similar to the ground truth.

\subsection{Qualitative Evaluation} \label{sec:qualitative}

We perform qualitative analyses to evaluate the seismological fidelity and reliability of HEGGS-generated waveforms.

\subsubsection{Waveform Analysis}\label{subsec:qualitative_waveform}
Synthetic waveforms can be visually inspected alongside real seismic waveforms for similarities in terms of waveform morphology, including amplitude, shape, and duration of seismic signals. 

\begin{figure}[t]
  \centering
  \includegraphics[width=\columnwidth]{images/results_all/ALL_wave_dpi300.pdf}
  \vspace{-0.55cm}
  \caption{Comparison of 3-component real and synthetic waveforms from earthquake datasets SCEDC (North America), KMA (East Asia), and INSTANCE (Europe).
  For each panel, top shows a real waveform and the bottom shows a synthetic waveform generated with the same metadata. 
  The phase arrivals marked as red (P) and blue (S) lines are detected by EQT. 
}\label{fig:waveform}
\end{figure}

\cref{fig:waveform} compares a set of representative synthesized waveforms and real waveforms from three datasets. 
Notably, both synthesized waveform and real waveform depict similar patterns of seismic activity, including distinct seismic phases and their corresponding arrivals. This alignment underscores the effectiveness of our approach in accurately replicating the seismic signal's morphology and temporal evolution. More waveform examples can be found in \cref{sec:appendix_waveform_spectrogram}.
\subsubsection{Spectrogram Comparison}\label{subsec:qualitative_spectrogram}

We also show the output spectrogram of HEGGS, compared to the spectrogram of the ground truth waveform to examine their time-frequency characteristics.
This provides insights into the similarities of temporal distribution of energy across different frequency bands. 
\begin{figure}[t]
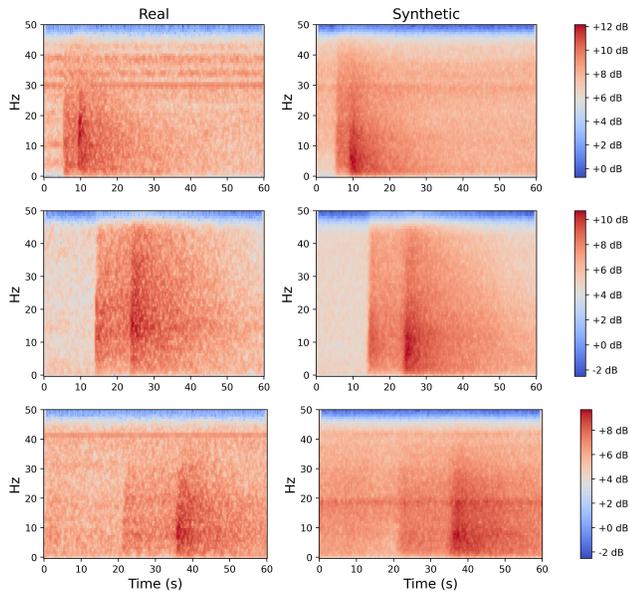

\begin{subfigure}{\linewidth}
  \centering
  \includegraphics[width=\columnwidth]{images/results_all/scedc_spec_dpi500.pdf}
  \vspace{-0.44cm}
  \caption{SCEDC}
  \vspace{0.14cm}
  \label{fig:spectrograms_scedc}
\end{subfigure}

\begin{subfigure}{\linewidth}
  \centering
  \includegraphics[width=\columnwidth]{images/results_all/kr_spec_dpi500.pdf}
  \vspace{-0.44cm}
  \caption{KMA}
  \vspace{0.14cm}
  \label{fig:spectrograms_kma}
\end{subfigure}
\begin{subfigure}{\linewidth}
  \centering
  \includegraphics[width=\columnwidth]{images/results_all/instance_spectrogram_38.pdf}
  \vspace{-0.44cm}
  \caption{INSTANCE}
  \label{fig:spectrograms_instance}
\end{subfigure}
\vspace{-0.7cm}
\caption{Comparison of real and synthetic spectrograms.}\label{fig:spectrograms}
\vspace{-0.5cm}
\end{figure}

In \cref{fig:spectrograms} we show the comparison of spectrograms, where each corresponds to the waveforms in \cref{fig:waveform}.
Upon comparing the synthesized spectrogram with the real spectrogram, several key observations come to light. 
Both spectrograms exhibit remarkable similarities in terms of phase arrival times and frequency band distribution, indicative of the efficacy of our synthesis approach in capturing essential seismic signal characteristics.
However, it is discernible that the synthesized spectrogram exhibits a slightly lower resolution compared to the real spectrogram, with some details appearing less defined. 
This reduction in resolution is particularly evident in the depiction of fine-scale frequency variations and subtle signal features. 
Despite this difference, the overall agreement between the synthesized and real spectrograms underscores the fidelity of our synthesis method in reproducing the fundamental characteristics of seismic signals. More spectrogram examples are shown in \cref{sec:appendix_waveform_spectrogram}.

\subsubsection{Frequency Content Analysis}\label{subsec:qualitative_freqcontest}

\begin{figure*}[th]
\begin{subfigure}{0.33\linewidth}
    \centering
    \includegraphics[width=\columnwidth]{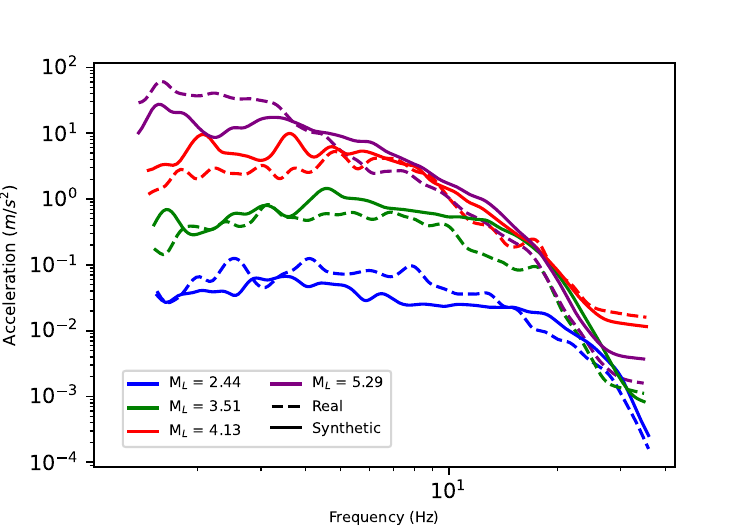}
   \vspace{-0.4cm}
    \caption{SCEDC}
   \label{fig:frequency_contents_scedc}
\end{subfigure}
\begin{subfigure}{0.33\linewidth}
    \centering
   \includegraphics[width=\columnwidth]{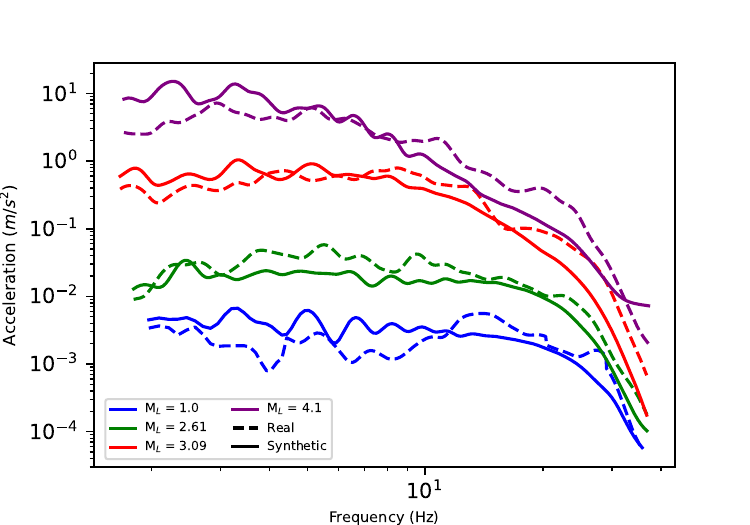}
   \vspace{-0.4cm}
   \caption{KMA}
   \label{fig:frequency_contents_kma}
\end{subfigure}
\begin{subfigure}{0.33\linewidth}
    \centering
   \includegraphics[width=\columnwidth]{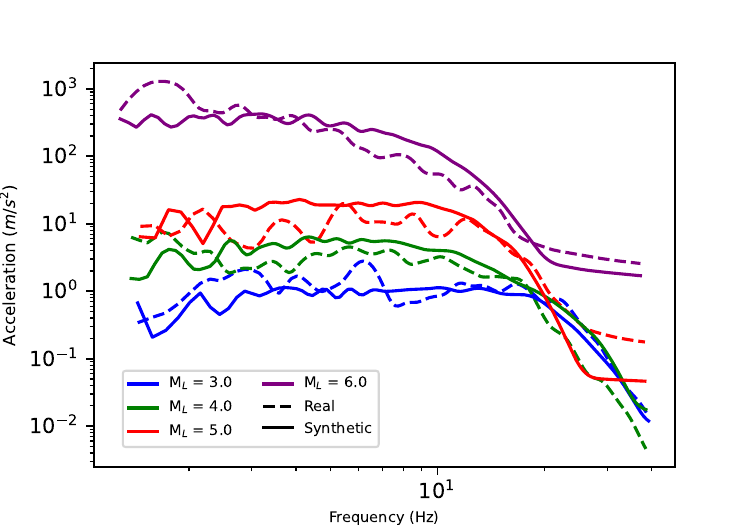}
   \vspace{-0.4cm}
   \caption{INSTANCE}
   \label{fig:frequency_contents_instance}
\end{subfigure}
\vspace{-0.7cm}
\caption{(a)-(c): Frequency contents of synthetic waveform compared to the real waveform}
\vspace{-0.65cm}
\label{fig:frequency_contents}
\end{figure*}

We also analyze how the energy released during an earthquake is retained in different frequencies. 
This analysis is closely related to the concept of corner frequency \cite{boore1983stochastic} and the seismic moment ($M_{0}$). The corner frequency is generally associated with the earthquake's magnitude. Specifically, the corner frequency identifies the point at which high-frequency energy begins to decline sharply, indicating that larger earthquakes generally have lower corner frequencies. The $M_{0}$ represents the total energy released by the earthquake, which corresponds to an increase in amplitude on the spectrum as the earthquake's magnitude increases. By comparing synthetic and observed seismic signals, we aim to evaluate the similarity between the two characteristics of corner frequency and $M_{0}$ across different magnitudes. 

We apply both Fast Fourier Transform (FFT) and Konno-Ohmachi-smoothing \cite{konno1998ground} technique to enhance our comparison. Also, we apply Wood-Anderson simulations \cite{Havskov2010} to compare results from distinct stations. 
The resulting spectra are shown in \cref{fig:frequency_contents_scedc,fig:frequency_contents_kma,fig:frequency_contents_instance}. We observe significant differences in corner frequency and $M_{0}$ across different earthquake magnitudes. 
We also observe the trend of reduced corner frequency reduces and increased $M_0$ as the magnitude grows.

\subsubsection{Magnitude Manipulation}\label{subsec:qualitative_magnitude}

\begin{figure}[t]
  \centering 
  \includegraphics[width=0.8\columnwidth]{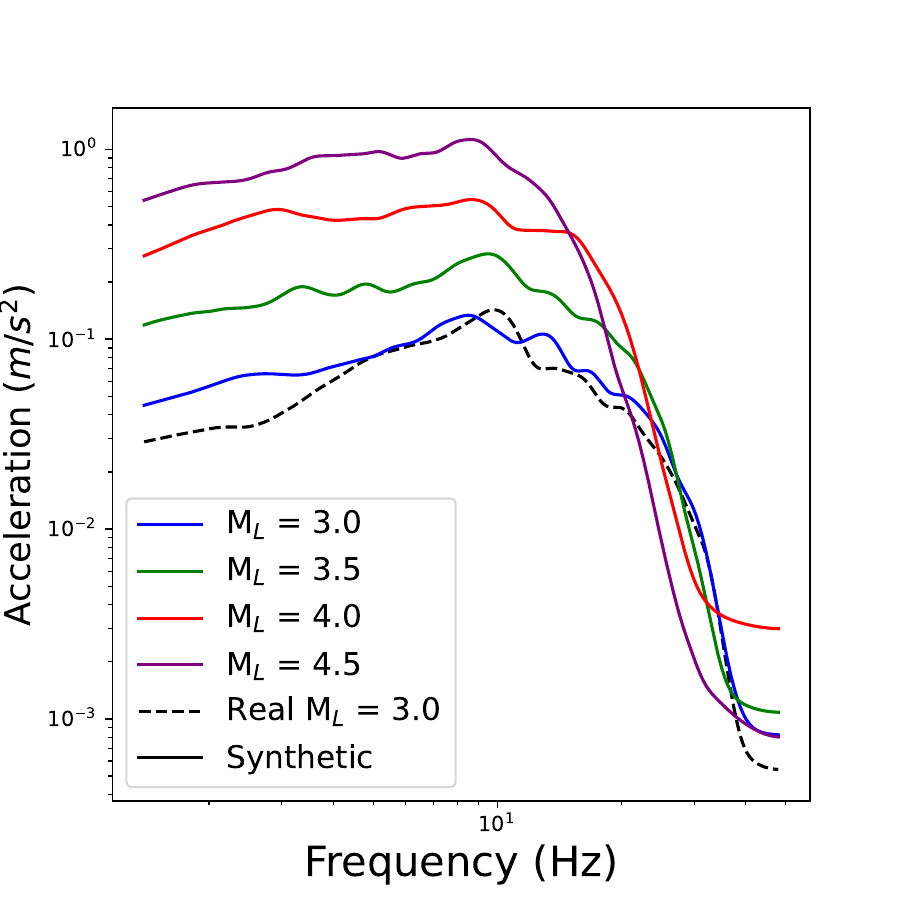}
  \vspace{-0.4cm}
  \caption{Magnitude Manipulation}\label{fig:frequency_contents_magnitude_manipulation}
\end{figure}

Synthesis of waveforms from fictitious earthquake is difficult challenging problem in DL-seismology area, especially with large magnitude, since the seismological features of large earthquake is hard to capture and large magnitude earthquake data is very rare, which requires the extrapolation ability of the model.
We select $c_{tgt}$ from the test dataset, change $M_L$, and generate waveform with modified $c_{tgt}'$, without $W^{src}$ and analyze the frequency contents in \cref{fig:frequency_contents_magnitude_manipulation}. The result is quite promising: the corner frequency gets smaller and $M_0$ gets larger properly when the magnitude grows, which is consistent to the theory as explained in \cite{source_spectra_fig}.

\subsubsection{Waveform Analysis on Synthetic Stations}\label{subsec:qualitative_sectionplot}

By arranging virtual observation stations in a linear manner, spatial variations of seismic waves could be observed, facilitating an understanding of seismic event characteristics. 
The synthesized seismic waves reflected seismic activity at the virtual observation stations, enabling exploration of subsurface structures and seismic wave propagation characteristics.

The sections in \cref{fig:section_plots} represented the positions of observation stations horizontally and represented the temporal and frequency characteristics of seismic activity vertically.
Through such visualization, comparisons between synthesized and observed seismic waves could be conducted, assessing the fidelity of the synthesized seismic waves in reflecting seismic events. 
Results from section plots clearly visualized spatial and temporal variations of seismic activity, serving as crucial criteria for evaluating the extent to which HEGGS accurately reproduces actual observed results. More section plot examples of seismic events are provided in \cref{sec:appendix_section_plot}.

\begin{figure}[t]
  \centering 
  \includegraphics[width=\columnwidth]{images/results_scedc_v2/synthetic/section_plot/37917624_section_plot_together_e_E_wo_synthetic_text.pdf}
  \vspace{-0.72cm}
  \caption{Section plots comparing synthetic and real waveforms. The vertical capital letters displayed at the top of the plot are the station ID that observed each real waveform.}\label{fig:section_plots}
\end{figure}

\section{Ablation Studies}
Compared to the conventional latent diffusion model, we introduced two major components, the efficient learning framework and amplitude correction module, to generate high-quality seismic waveforms. In this section, we present the results of the ablation study to evaluate the role of each component, on SCEDC dataset.

To assess the effectiveness of this approach, we conduct training of the LDM \cite{LDM} with two distinct training schemes: the original and modified one trained by \cref{eqn:loss-ldm} with paired data.
Unfortunately, conventional LDM training on our dataset was diverged. Hence we tried to train LDM to generate normalized waveform, which is a relaxed version of our task. 

After that, we tried to generate unnormalized waveforms, by changing the training framework. Preserving the model architecture, we trained the model which has same architecture, but by end-to-end training \cref{eqn:loss-ours}. The only difference between this model and ours is the amplitude correction module ACM. 

\begin{table} 
\caption{Results of ablation study. $^*$ represents the generation of normalized waveform. $env.corr$ refers to the envelope correlation between synthesized waveform and real waveform.}
\vspace{0.26cm}
  \label{tab:ablation}
  \centering
  \begin{tabular}{c@{\hskip3pt}c@{\hskip3pt}c@{\hskip3pt}c}
    \toprule
    Model  &P\_MAE (s)   & S\_MAE (s) & $env. corr$ \\
    \midrule
    LDM$^*$ & 1.1142   & 1.7294  & 0.6932 \\
    +paired data$^*$&  0.5633  & 0.7808  & 0.7726  \\
    +end-to-end train&0.8014&1.5367&0.6239\\
    +ACM (HEGGS)&0.4760&0.5476&0.8187\\
    \midrule
    LDM+ACM&1.1131&1.6372&0.6981\\
    +paired data&0.7748&0.9402&0.7965\\
    \bottomrule
  \end{tabular}
\end{table}

The results can be found in \cref{tab:ablation}. On first two rows, learning with paired data were very effective to increase the quality of waveform, especially as the phase arrival times were twice as accurate.
Comparing 2nd and 3rd rows, the overall scores seem to be worse, but the model in 2nd row often generates unrealistic waveforms in qualitative analysis results in \cref{sec:appendix_evidence}.
Also, note that the result of 3rd row is the result of unnormalized waveform generation while 2nd row generates normalized waveform.
Even the difficulty of generation problems were increased, the paired training shows better results, compared to the baseline model LDM. This may indicate the failure of VAE pretraining that pretrained VAE could not capture the amplitude as important feature. 
The amplitude correction module ACM helps to improve the quality of seismic waveform synthesis, as shown in the 3rd and 4th rows of \cref{tab:ablation}. 

Thanks to reviewers, we found that the ACM has ability to allow LDM trainable with unnormalized waveforms, as shown in 5th row of \cref{tab:ablation}. 
The last row of \cref{tab:ablation} shows the remarkable improvement induced by pair-exploiting strategy, but still not better than HEGGS with end-to-end training.

\section{Discussion}
The HEGGS training method, which takes advantage of the seismic dataset characteristic by training the model with paired data, allows for two modes of generation criteria: with and without $W^{src}$. Although HEGGS is trained using $W^{src}$, the fidelity of generation without $W^{src}$ is promising, and much better than the benchmark models.
Generation without $W^{src}$ allows us to synthesize waveforms of non-existent earthquakes and simulate the ground motion with different magnitude or location, which is big challenge in seismology. As shown in \cref{subsec:qualitative_magnitude}, HEGGS shows the theoretically-expected trend of corner frequency and $M_0$, but may not be perfect since we only used minimal condition about location and magnitude.
We expect larger success with additional geological features, which we did not included in minimal condition, for this non-existent earthquake synthesis challenge.

The seismic synthesis studies are inevitably build on regional dataset, since each observatories are operated independently by each government and thus the waveform formats are unaligned. Another challenging problem arises here to build global model by training multiple models on individual dataset, with consideration of robust consistency, especially on border. We expect HEGGS would be the effective starting point of this research direction.

Compared to other methods, HEGGS shows superior fidelity with minimal conditions, especially for the P/S phase arrival times. We expect HEGGS would applicable to downstream tasks which is sensitive to the phase arrival times,such as early warning systems, earthquake modeling and disaster, hence we are planning to develop algorithms for those downstream tasks using HEGGS as a near-future research.

\section{Conclusion}

In this paper, we propose HEGGS, an efficient training framework for seismic waveform synthesis utilizing a diffusion model and a minimal set of conditions. Our approach generates seismic waveforms using only readily accessible information, such as location and magnitude, thereby avoiding the need for extra conditions.

To empirically validate the proposed method, we constructed a seismic dataset from the SCEDC, INSTANCE and KMA dataset by collecting simultaneously paired observations aligned with the earthquake's origin time.
We demonstrate that HEGGS produces more realistic waveforms than existing benchmark models by applying seismic domain-specific metrics, such as envelope correlation and P/S phase arrival times, for expert-level comparison and applications.

\section*{Impact statement}
Our work enables high-fidelity seismic waveform synthesis, enhancing earthquake modeling, early warning systems, and disaster preparedness while promoting AI use in geophysical research.

\section*{Acknowledgements} 
The authors would like to thank Prof. Seongryong Kim and Dr. Jongwon Han of Seismology Lab, Korea University for their professional opinions and feedback on the seismological aspects of this work. 
The authors are supported in part by the Korea Meteorological Administration Research and Development Program under
Grant KMI2021-01112 (RS-2021-KM211112).

\bibliography{aimlk}
\bibliographystyle{icml2025}

\newpage
\appendix
\onecolumn
\section{Dataset Construction}\label{sec:appendix-dataset}

We used three datasets (SCEDC\cite{scedc}, KMA\cite{h_magnitude} and INSTANCE\cite{instance}) from different regions. In this section, we explain how each dataset was constructed. All datasets are collected from corresponding APIs and processed to have 60-seconds duration and applied $1\sim45\text{Hz}$ bandpass filter. 

We split each dataset into training dataset and test dataset, according to the earthquake event, to evaluate the fidelity of generated waveform for the earthquake which is unseen during the training. 

\begin{table}[h]
    \centering
        \caption{Features of each dataset}
        \begin{tabular}{ccccccc}
        \toprule
        \multicolumn{1}{c}{dataset} & \multicolumn{2}{c}{SCEDC} & \multicolumn{2}{c}{KMA} & \multicolumn{2}{c}{INSTANCE} \\
        Features & Train & Test & Train & Test & Train & Test \\
        \midrule
        \#observations & 71,488 & 17,878 &  237,755 & 58,925 & 72,904 & 19,872 \\
        \#source event & 2,098 & 525 &  2,052& 514 & 2,265 & 593 \\
        \#station & 149 & 149 & 134 & 134 & 578 & 534 \\
        average \#station per events & 34.07 & 34.05 & 115.87 & 114.64 & 24.43 & 25.29 \\
        average magnitude & 2.45 & 2.45 & 1.45 & 1.45 & 3.36 & 3.36 \\
        average epicentral distance & 125.25 & 126.71 & 235.48 & 234.22 & 57.82 & 57.79 \\
        average focus depth & 8.51 & 8.65 & 11.52 & 11.73 & 12.47 & 11.97 \\ 
        \bottomrule
        \end{tabular}
    \label{tab:scedc_dataset}
\end{table}

\subsection{SCEDC}
We exploit earthquake catalog of SCEDC \cite{scedc} provided by SeisBench\cite{seisbench}. We selected waveforms with a sampling rate of 100Hz that included 60 seconds from the earthquake and applied a bandpass filter in the $1\sim45\text{Hz}$ range to construct our data. Unfortunately, the Seisbench-provided dataset had fewer than 13 stations per earthquake events on average, therefore we utilized Obspy API\cite{beyreuther2010obspy} to collect additional observations on more stations in the station list of \cite{scedc_mag} for each earthquake.
Using earthquakes from the catalog during the years 2016 to 2019, we constructed a new dataset with approximately 34 stations per source. The \cref{tab:scedc_dataset} shows the count of datasets we used.

The $V_{S30}$ information was sourced from \cite{USVs30} and used only during the GMPE analysis, not during the training or model inference processes. The average value was used if multiple $V_{S30}$ values were present for a single station code. For station codes without $V_{S30}$ data, $760 m/s$ was assigned to negate the influence of $V_{S30}$ during GMPE analysis.

\subsection{KMA}
KMA data source consist of continuous waveform data were employed, which are operated by KMA (Korea Meteorological Administration) and KIGAM (Korea Institute of Geoscience and Mineral Resources). We exploit the dataset appear in \cite{h_magnitude} which is constructed from earthquake catalog provided by KMA, spanning from 2016-2020, and used subset consist of observations from broadband sensors. 
Similarly to SCEDC, the waveforms have a sampling rate of 100hz, a duration of 60 seconds, and a frequency $1\sim45\text{Hz}$.

\subsection{INSTANCE}

We used the Seisbench-provided version INSTANCE dataset and created a subset by selecting only the traces satisfying:
\begin{enumerate}
    \item includes records for 60 seconds from the earthquake occurrence time
    \item local magnitude is larger than 3.0.
    \item P-arrival time is included in the metadata to ensure that the earthquake signal is observed.
\end{enumerate}

For the EQT evaluation in \cref{tab:EQT}, we excluded waveforms which include multiple event signals, which are out of our scope.

\section{Implementation details}\label{sec:appendix-implementation}
We implement the proposed model with following implementation details.

During the training, $W_{src}$ is fixed for a specific earthquake source ID, and $W_{tgt}$ is sampled from earthquakes with the same source ID. Among these, if metadata contained P/S phase labels, samples are randomly selected from those with labels. If P/S phase labels are absent, samples are chosen randomly without considering P/S phase labels. And also we conduct preprocessing of seismic data.

We implement using single NVIDIA-RTX A6000 with 48GB memory. For training, we set the number of epochs to 500 and the training batch size to 4. To enhance training efficiency, we apply an accumulation step 4, resulting in an effective batch size of 16.
For the loss, we set the maximum diffusion steps to $T=1000$ and SNR weight 5. We minimize the loss by AdamW optimizer with learning rate $10^{-5}$ and \emph{pytorch.optim} defaults. During the training, we applied learning rate decaying technique with linear scheduler. The total duration of training is approximately 65 hours. 
\subsection{Neural Network Architecture}\label{subsec:appendix_neuralnet}
We utilize the U-Net backbone with cross-attention architecture similar to  \cite{LDM, tango}, to represent $\mathbf{m}_\theta$, with modification in the domain-specific encoder $\tau_\theta$ to map $\vec{c}_{tgt}$ to hidden feature $\tau_\theta(\vec{c}_{tgt})$. For the implementation, we construct $\tau_\theta$ by 5-layer FFN model. The encoded conditional vector $\tau_\theta(\Vec{c}_{tgt})$ will be provided as a value and key of cross attention module $Attn(Q,K,V)$ while U-Net feature is provided as query $Q$.

For $\mathcal{E}_{AE}$ and $\mathcal{D}_{AE}$, 
we take same architectures from VAE of \cite{VAE} and give a modification on $\mathcal{D}_{AE}$. With the vanilla module $\mathcal{D}_{AE}$, we find that the proposed model is not effective in accurately predicting the amplitude of the output waveform. 
Therefore, we propose to attach an additional module ACM after $\mathcal{D}_{AE}$ to predict the amplitude correction feature and multiply it to the predicted spectrogram. In detail, we utilize the encoder, TSConformer blocks and Magnitude mask decoder module from MP-SeNet \cite{lu2023mp} and provide output of $\mathcal{D}_{AE}$ and auxiliary phase spectrogram induced by GriffinLim algorithm to correct the amplitude and enhance the quality of generation. Improving the original implementation  \cite{lu2023mp} that allows only reducing the output, we add four TSConformer blocks and replace the final sigmoid activation function with Softplus function to provide the ability to increase as well.

\section{Pre-processing Recipe}\label{sec:appendix-preproc}

\subsection{Conditional Vector Pre-processing} 
We explain the process of $\vec{c}_{tgt}$ constuction. Recall the variables that we are used to synthesize waveform are:

\begin{enumerate}
    \item $s_{lat}, s_{lon}$ : latitude and longitude of the station to observe the waveform data. 
    \item $e_{lat}, e_{lon}$ : latitude and longitude of epicenter.
    \item $e_{dep}$ : depth of the hypocenter, unit of kilometers.
    \item $M_L$ : magnitude of the earthquake.
\end{enumerate}

We preprocessed those variables to construct an 11-dimensional condition vector and later provide it to our condition encoder module $\tau_\theta$. 

First of all, we encode locational information $s_{lat},s_{lon},e_{lat}$ and $e_{lon}$ with the following process:

\begin{enumerate}
\addtolength{\itemindent}{1cm}
    \item Normalize the values to get $s_{lat}',s_{lon}',e_{lat}'$ and $e_{lon}'$ with following:
    \begin{equation}
        s_{lat}' = \frac{s_{lat}-l_{lat}}{u_{lat}-l_{lat}}, e_{lat}' = \frac{e_{lat}-l_{lat}}{u_{lat}-l_{lat}}, s_{lon}' = \frac{s_{lon}-l_{lon}}{u_{lon}-l_{lon}} \mbox{ and } e_{lon}' = \frac{e_{lon}-l_{lon}}{u_{lon}-l_{lon}}
    \end{equation}
    where $(l_{lat},u_{lat})$ and $(l_{lon},u_{lon})$ represent the lower and upper bounds of latitude and longitude, respectively, for the region of interest. 

    In our datasets, we summarize those bounds in \cref{tab:dataset_region}.

    \begin{table}[H]
    \caption{upper and lower bounds of the region of interest}
    \centering
    \begin{tabular}{ccccc}
        \toprule
        Dataset (region) & $l_{lat}$ & $u_{lat}$  & $l_{lon}$ & $u_{lon}$ \\
        \midrule
        SCEDC (Southern California) & 32.0& 37.9& -121.0 &-114.1\\
        KR (South Korea) &33.12&38.60&124.64&131.87\\
        INSTANCE (Italy) &35.00&48.03&5.32&20.01\\
        \bottomrule
    \end{tabular}
    \label{tab:dataset_region}
\end{table}

    \item Motivated from polar coordinate transformation\cite{Coordinate_Transformations}, which is commonly used in GPS field, we further encode normalized coordinate to following:
    \begin{equation}
        \begin{aligned}
            &c_{sta} = (cos(s_{lat}')cos(s_{lon}'), sin(s_{lat}')cos(s_{lon}'),sin(s_{lon}'))\\
            &c_{epi} = (cos(e_{lat}')cos(e_{lon}'), sin(e_{lat}')cos(e_{lon}'),sin(e_{lon}'))
        \end{aligned}
    \end{equation}
\end{enumerate}

Secondly, we compute the back azimuth angle $Azi$ and encode by
\begin{equation}
    c_{azi} = (cos(Azi), sin(Azi))
\end{equation}

Lastly, we compute and normalized epicentral distance $R_{epi}$, focus depth $d_s$ and magnitude $M_L$. Each are normalized by following formula:

\begin{table}[h]
\centering
\begin{tabular}{cccc}
\toprule
           & SCEDC & KMA & INSTANCE  \\
           \midrule
$R_{epi}'$ & $(R_{epi}-125.542401)/55.810322$&  $(R_{epi}-219.91)/119.99$&$(R_{epi}-57.8158)/31.7465$\\
$d_s'$ &$(d_s-8.564146)/4.658161$ &$(d_s-11.59)/5.40$ &$(d_s-12.3680)/13.2456$  \\
$M_L'$ &$(M_L-2.0)/6.4$     & $(M_L-0.35)/5.24$    & $(M_L-3.0)/6.5$    \\
\bottomrule
\end{tabular}
\end{table}
Concatenating the processed features $c_{sta},c_{epi},c_{azi},R_{epi}',d_s'$ and $M_L'$, we get an 11-dimensional conditional vector $\vec{c}_{tgt}$ for our problem, the synthesis of seismic ground motion. 

\subsection{spectrogram construction}

The generation target of out model is spectrogram, which is in time-frequency domain. We report the process of spectrogram construction as pre-processing. We employed the STFT (Short-Time Fourier Transform) with a hop length 16. Given that the spectrogram's scale is closely related to the earthquake's amplitude, we used an $nfft$ and $window~length$ of $128$ and applied a logarithmic scale transformation for better scale adjustment. Consequently, the original waveform data of size $3\times6000$ was reshaped into $3\times64\times376$.

\section{EQT Training Details}\label{sec:appendix-eqt}
We used EQTransformer \cite{mousavi2020earthquake} provided by SeisBench \cite{seisbench}. Starting from pre-trained model provided by SeisBench, we finetune the model with our dataset, with the same training protocol. After standardizing the waveforms, we trained the model using the Adam optimizer, with a batch size of 512 and a learning rate of $10^{-3}$, for 100 epochs. Other hyperparameters of the optimizer were set to default. 
For hyperparameter search, the learning rate ranged from $10^{-2}$ to $10^{-5}$, and the performance was best when it was $10^{-3}$.

\section{Details on Benchmark Models}\label{sec:appendix-benchmark}

\subsection{SeismoGen \cite{wang2021seismogen}}
SeismoGen is a CGAN-based model that generates waveforms conditioned on the presence of seismic events (e.g., P or S waves). The Discriminator takes both the waveform and the presence of seismic events as inputs. It then divides the waveform into high and low frequency components, analyzing each to determine if waveform is real or synthetic.
SeismoGen used data from three stations in Oklahoma: V34A, V35A, and V36A, while we used data from 149 stations from SCEDC. Our synthesis approach used station and earthquake information instead of presence of seismic events.
SeismoGen generated waveforms as 40 seconds at 40Hz, but we aimed for 60 seconds at 100Hz. We used an input noise length of 1500 and added upsampling at the end of the first convolution layer. The basic training used noise as input, and for comparison with HEGGS, we also trained using waveform. When using waveforms, we modified each pipeline to utilize one ENZ channel.
The hyper-parameters we used included the Generator learning rate and Discriminator learning rate are set to $10^{-4}$ and $10^{-6}$, using the RMSprop optimizer over 3000 epochs. The $\lambda$ is set to 10 when using noise and 15 when using the input waveform. We saved the best model based on envelope correlation. 
We experimented with learning rates ranging from $10^{-4}$ to $10^{-7}$, using both Adam and RMSprop optimizers. The value of $\lambda$ was tested at 5, 10, and 15. The best-performing combination of these parameters was selected for the final model. Additionally, the results reported in \cref{tab:quantitative_result} reflect the best performance achieved across 30 iterations.
Addressing the instability of the original method, we added the L1 loss \cref{eqn:loss-pix2pix} from pix2pix\cite{pix2pix} as an additional loss term to improve training stability.

\subsection{ConSeisGen \cite{conseisgen}}
ConSeisGen is an ACGAN-based model that generates waveforms conditioned on the epicentral distance. The Discriminator consists of two components: $D_P$, which learn determining whether the waveform is real or synthetic, and $D_Q$, which learn regression estimating the distance between the epicenter and the station. While ConSeisGen generated waveforms with 3 channels and a length of 4096, we aimed to generate waveforms with 3 channels and a length of 6000. We modified the first linear layer and removed upsampling in the final layer.
ConSeisGen used KiK-net data, which began recording shortly before the arrival of the P-wave. However, the SCEDC data utilized in this model was recorded from the onset of the earthquake for a duration of 60 seconds.
ConSeisGen generates waveforms based on the epicentral distance. However, waveforms can vary even at the same distance due to factors like magnitude and geological conditions. To generate waveforms for specific locations, we utilized minimal additional condition such as station data and source data along with the epicentral distance.
The hyper-parameters we used included the Generator learning rate and Discriminator learning rate are set to $2\times 10^{-4}$ and $10^{-5}$, using the Adam optimizer over 5000 epochs.
Referring eq.4 of \cite{conseisgen}, the loss function consists of Adversarial Loss, Regression Loss($L_{reg}$), and Diversity Improvement Loss($L_{di}$). The $L_{reg}$ computes the $l$1 loss between $D_Q$'s output and the condition vector, with the $\lambda_{reg}$ set to 1. The $L_{di}$ aims to prevent mode collapse by maximizing the distance between feature maps, with $\lambda_{di}$ set to 10 when using noise and 5 when using waveforms. 
We experimented with learning rates ranging from $10^{-4}$ to $10^{-6}$, using both Adam and RMSprop optimizers. The value of $\lambda_{di}$ was tested at 5, 10, and 15, while $\lambda_{reg}$ was fixed at 1. The best-performing combination of these parameters was selected for the final model. Additionally, the results reported in \cref{tab:quantitative_result} reflect the best performance achieved across 30 iterations.
Addressing the instability of the original method, we added the L1 loss \cref{eqn:loss-pix2pix} from pix2pix\cite{pix2pix} as an additional loss term to improve training stability.

\begin{equation}\label{eqn:loss-pix2pix}
    L_{\text{L1}}(G) = \mathbb{E}_{x,y,z} \left[ \left\| x_{tgt} - G(z,y) \right\|_1 \right]
\end{equation}

\subsection{BBGAN \cite{BBGAN}}
BBGAN is a conditional generative model within the Wasserstein GAN framework.
The original conditions of BBGAN are $V_{S30}$, earthquake magnitude, and epicentral distance. We modified conditional vector to ours, add conditional vector encoder $\tau_\theta$ to both generator and discriminator, modified the last upsample layer of generator to have scale factor 3 (original: 2), and lastly increased the number of hidden features of last convolution block of discriminator, corresponding to our waveform shape $(3,6000)$. Those changes allows the model to generate $(3,6000)$ shape waveform from the provided conditional vector. To further improve the performance, we replaced all relu activations of generator and leaky relu activations of discriminator to gelu activation. 
Additionally, while the original BBGAN paper utilized data from Japanese networks K-NET and KiK-net with earthquake magnitudes larger than 4.5, our approach employed data from the SCEDC \cite{scedc} with earthquake magnitude larger than 2.0 for training.
In the training process, we set 500 training epoch and batch size 32, and Adam optimizer with learning rate $5\times10^{-7}$ and $\beta=(0.9, 0.999)$. Also the final loss function is composed of adversarial loss, L1 reconstruction loss, and a KL divergence term. The L1 regularization term was set to 25, and the KL regularization term was set to 0.01. For evaluation during the validation loop, envelope correlation was used as the performance metric. During the training, the linear learning rate decay technique was applied.

\subsection{LDM \cite{LDM}}
\subsubsection{VAE \cite{VAE} pretraining} \label{subsec:VAE}
Due to lack of pretrained weights of VAE trained on seismic spectrogram, we first need to train VAE to encode $X^{tgt}$ and $X^{src}$ to latent vector $Z^{tgt}$ and $Z^{src}$.

Employing equation (25) of \cite{LDM}, we set the loss function for VAE training is:

\begin{equation}
    L_{total} = min_{\mathcal{E}_{AE}, \mathcal{D}_{AE}}, max_{\psi }[L_{rec}(x, \mathcal{D}_{AE}((x))) - L_{adv}(\mathcal{D}_{AE}(\mathcal{E}_{AE}(x))) + logD_{\psi}(x) + \lambda_{kl}KL]
\end{equation}
where $\lambda_{kl}$ is low weighted Kullback-Libler regularization term by factor $10^{-6}$.

Unfortunately, the VAE training on our spectrogram diverged, due to difficulty on magnitude processing. Therefore, we apply standardization on spectrogram to relax the problem. And the latent space size is $64\times16\times94$.

We report reconstruction performance of the Auto-encoder model using the proposed our 
metrics. The reconstruction performance results as follow in \cref{tab:vae_result}.

\begin{table}[H]
  \caption{Reconstruction result}
  \centering
  \begin{tabular}{ccccccc}
    \toprule
        &\multicolumn{5}{c}{waveform} &\multicolumn{1}{c}{spectrogram}\\
    \cmidrule(r){2-6}\cmidrule(r){7-7}
 Model& P\_MAE (s) & S\_MAE (s) & envelope corr & SNR & PSNR & MSE \\
 \midrule
 VAE& 0.5155 & 0.7066 &  0.7567 & -2.9984 & 25.1800  & 0.2459\\
    \bottomrule
  \end{tabular}\label{tab:vae_result}
\end{table}

\subsubsection{LDM \cite{LDM}} \label{sec:Detail LDM training}
We train LDM using the pretrained VAE \cref{subsec:VAE} and DDPM\cite{DDPM} scheduler. Additionally, the overall model architecture is adapted and modified base on the TANGO \cite{tango} model and code. But, while TANGO models incorporate text-encoded conditions through Large Language Model, the seismic data does not exist text conditions. Therefore, we employ our preprocessed conditions 
and apply our conditional vector encoder $\tau_\theta$
for training. During model training, the learning target is set the samples from the DDPM scheduler. Training is conduct using two methods and training losses.
\begin{itemize}
    \item \cref{eqn:loss-ldm1}: not utilizing the characteristic of paired data
    \item \cref{eqn:loss-ldm2}: utilizing the characteristic of paired data
\end{itemize}
We set the hyperparameters for the AdamW optimizer as follows: an initial learning rate $10^{-5}$ and $\beta=(0.9, 0.999)$, and a weight decay of $10^{-2}$ and adam epsilon $10^{-8}$. Also, we apply the learning rate decaying technique
with the linear scheduler. The training batch size is set to 4 with an accumulation step of 4, resulting in a total effective batch size of 16. The model is trained for 500 epochs. The results 
indicate that training with paired data outperforms training without paired data.
\begin{equation}\label{eqn:loss-ldm1}
    L_{LDM}' = \mathbb{E}_{(Z^{tgt},\Vec{c}_{tgt}),\epsilon,t} \|Z^{tgt}-\mathbf{x}_\theta(z_t^{tgt},\Vec{c}_{tgt},t)\|
\end{equation}

\begin{equation}\label{eqn:loss-ldm2}
    L_{LDM}' = \mathbb{E}_{(Z^{src}, Z^{tgt},\Vec{c}_{tgt}),\epsilon,t} \|Z^{tgt}-\mathbf{m}_\theta(z_t^{src},\Vec{c}_{tgt},t)\|
\end{equation}

\section{GMPE formula}\label{sec:appendix_gmpe}
In this section, we express the PGA formula for GMPE analysis. Given waveform $W$, we obtain local magnitude $M_L$ first, and compute the PGA value later.

\subsection{SCEDC\cite{scedc}}
Given the waveform $W$, the local magnitude $M_L$ of SCEDC can be computed by using the following formula (equations 1 to 6 of \cite{scedc_mag}) 
\begin{equation}
    \begin{aligned}
        M_L &= log A - log A_0(R_{hypo}) \mbox{ where }\\
        -log A_0(R_{hypo}) &= 1.11logR_{hypo} + 0.00189\times R_{hypo} + 0.591 + \sum^6_{n=1} \text{TP}(n) \times T(n,z).
    \end{aligned}
\end{equation}

where $A$ is amplitude of $W$ and $A_0(r)$ is attenuation function of southern california region. The station adjustment term was not applied due to a lack of values for recently installed stations.

The $\text{TP}(n)$ coefficients are

\begin{equation}
    \begin{aligned}
        \text{TP}(1) = +0.056,&\quad \text{TP}(2) = -0.031, \\
        \text{TP}(3) = -0.053,&\quad \text{TP}(4) = +0.080, \\
        \text{TP}(5) = -0.028,&\quad \text{TP}(6) = +0.015,
    \end{aligned}
\end{equation}

When $z$ is 

\begin{equation}
    z(r) = 1.11366 \times log(r) - 2.00574,
\end{equation}

$8 \leq r \leq 500$ to $-1 \leq z \leq +1$, $T(n,z)$ is the Chebyshev polynomial

\begin{equation}
    T(n,z) = cos[n \times \arccos(z)].
\end{equation}

After determining the local magnitude $M_L$, we obtain the PGA value by equation 1 of  \cite{scedc_gmpe}, with pynga \citep{pynga} implementation. Since HEGGS doesn't exploit the focal mechanism information, we set $mech$ and $rake$ to be 0, which represents unspecified. 
\subsection{KMA\cite{kr_gmpe}}
For the KMA dataset, the local magnitude $M_L$ can be computed by the following equation (equations 1 and 6 of  \cite{s_magnitude}):

\begin{equation}
    \begin{aligned}
        &M_L = log A - log A_0 + S\\
        &-log A_0 = 0.5869log(R_{epi}/100) + 0.001680(R_{epi}-100) + 3
    \end{aligned}
\end{equation}
where $A$ is the peak amplitude of the Wood-Anderson simulated waveform and $S$ is station-wise correction term and $R_{epi}$ is epicentral distance in kilometers.

After determining local magnitude $M_L$ we obtain PGA value $Y$ by  \cite{kr_gmpe} with following formula for South Korea peninsula:
\begin{equation}
\begin{aligned}
        log Y = &-3.16 + 0.75M_L \\&-0.72log\left[\sqrt{R_{epi}^2+3.7^2}\right] -0.0034R_{epi}
\end{aligned}
\end{equation}

\subsection{INSTANCE\cite{instance_gmpe_a,instance_gmpe_b}}
the local magnitude $M_L$ on the INSTANCE dataset can be computed by the following equation (equation 1 to 14 of \cite{inatance_mag}):
\begin{equation}
    \begin{aligned}
        M_L &= log A - log A_0(R_{hypo}) + C\\
        &= log A + 1.749log(R_{hypo}/100) + 0.0016(R_{hypo} - 100) + 2.9445 + C
    \end{aligned}
\end{equation}
where $A$ is the peak amplitude of the Wood-Anderson simulated waveform and $C$ is the station-wise correction term.

After determining the local magnitude $M_L$, we obtain the PGA value by equations 1 to 5 of \cite{instance_gmpe_b} and 7 to 8 of \cite{instance_gmpe_a}. In these equations, Moment Magnitude($M_W$) was used to compute PGA. However, \cite{inatance_mag} proposed a formula that satisfies $ M_L = M_W $, on average. Therefore, this study used $ M_L $ instead of $ M_W $.
Also, HEGGS doesn't exploit the focal mechanism information, we set the style of faulting $SOF$ to 0, representing the normal fault type. 

\section{Qualitative Analysis on Ablation Models} \label{sec:appendix_evidence}

This section is dedicated to the qualitative analysis of the SCEDC of the models mentioned in \cref{tab:ablation}. The figure compares the Real observation (Real), HEGGS(w/ ACM), end-to-end train (w/o ACM), and LDM + paired data. The human-labeled P/S arrival times of the earthquake are indicated by orange and black lines, while the P/S arrival times detected by EQT for each waveform are shown in red and blue.

\subsection{Positive samples}
We first list the positive samples, which are the results that all models generated realistic waveforms with accurate phase arrivals.

\begin{figure}[H]
    \centering
    \includegraphics[width=0.85\textwidth]{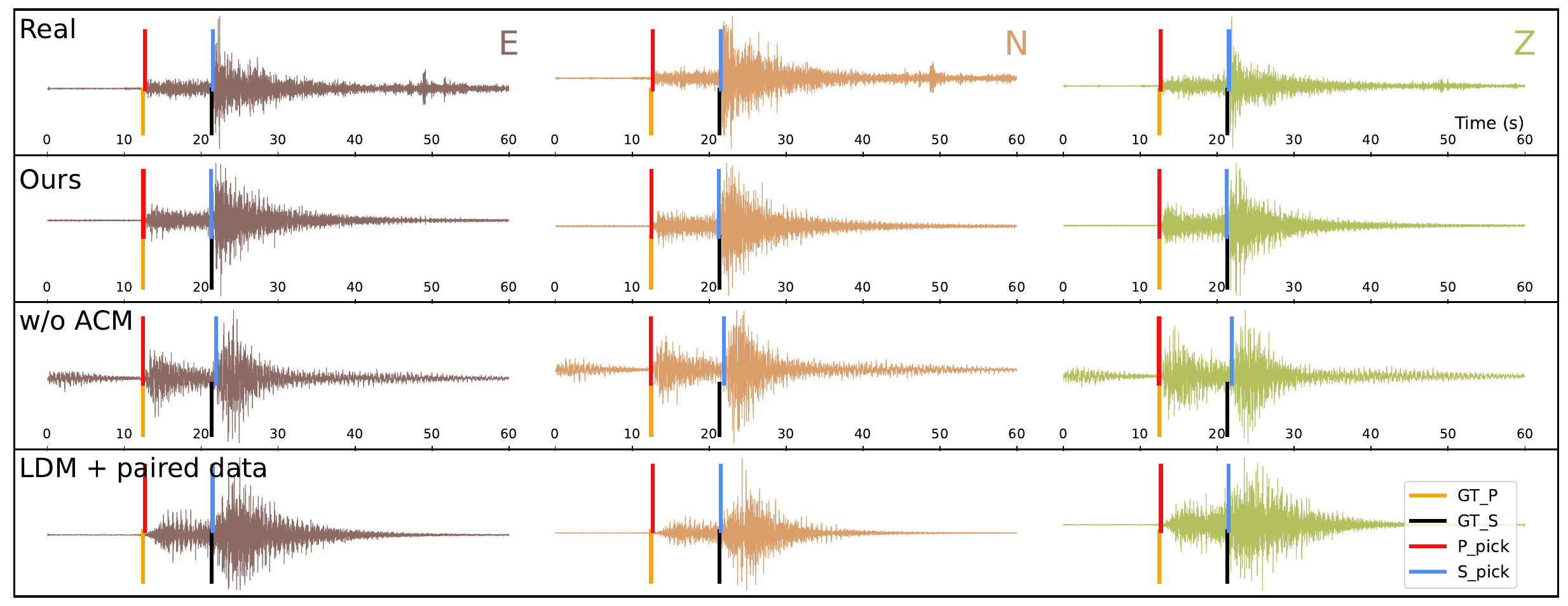}
    \caption{Positive synthesis results of our model and ablation models, compared to the real observation.}
    \label{fig:ablation_2}
\end{figure}

\begin{figure}[H]
    \centering
    \includegraphics[width=0.85\textwidth]{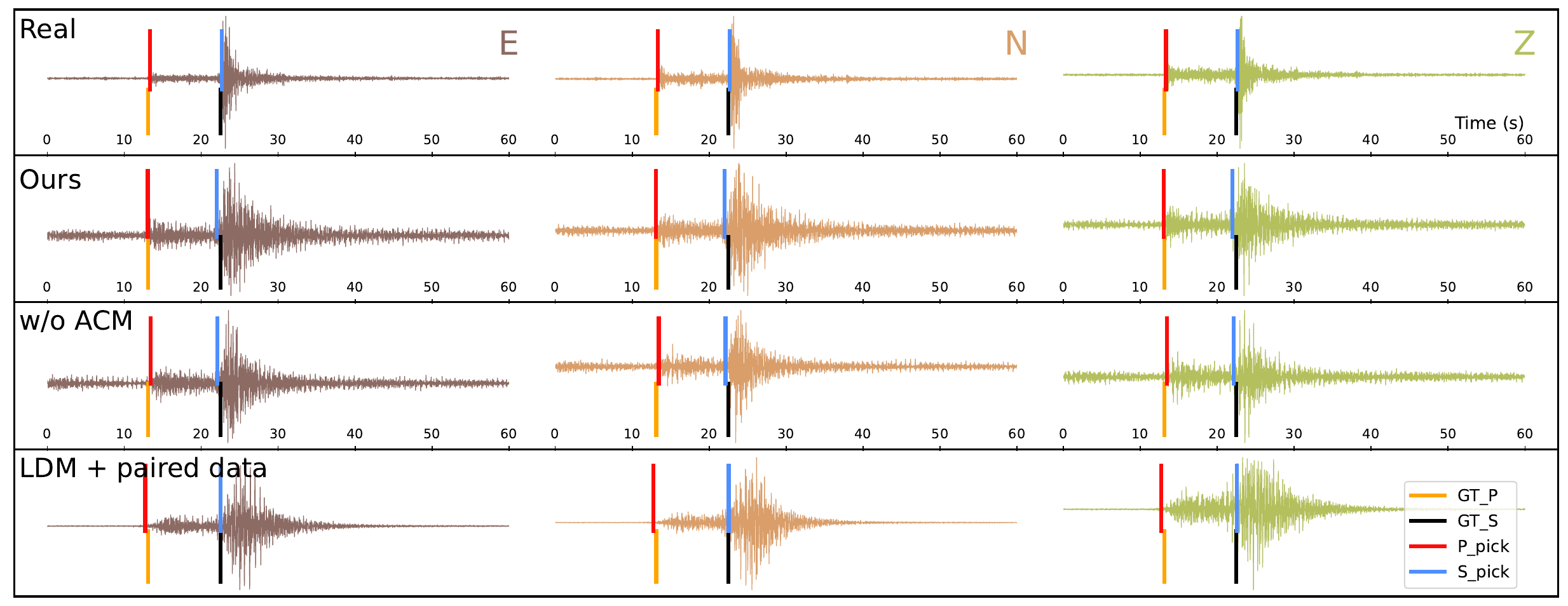}
    \caption{Positive synthesis results of our model and ablation models, compared to the real observation.}
    \label{fig:ablation_68}
\end{figure}

\begin{figure}[H]
    \centering
    \includegraphics[width=0.85\textwidth]{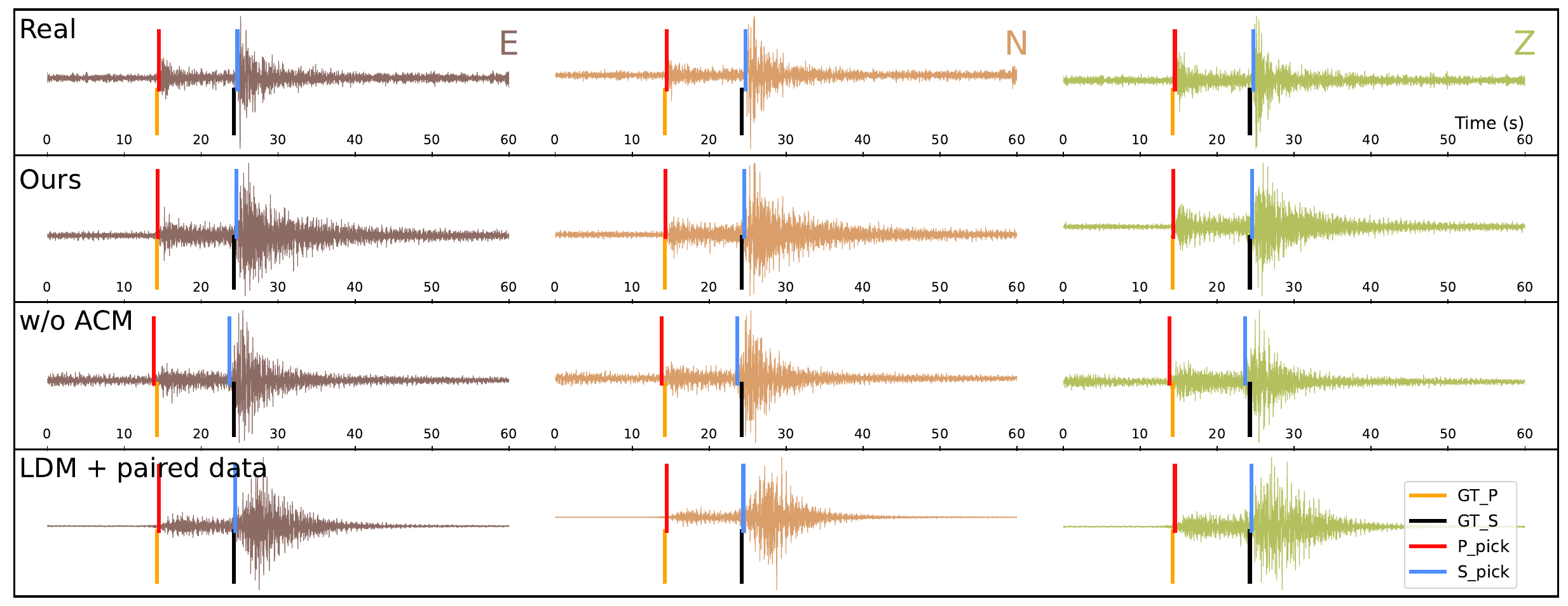}
    \caption{Positive synthesis results of our model and ablation models, compared to the real observation.}
    \label{fig:ablation_8}
\end{figure}

\subsection{Negative Samples}
We also include the results of the synthesis that at least one of the models failed to generate realistic and accurate waveforms.

\begin{figure}[H]
    \centering
    \includegraphics[width=0.85\textwidth]{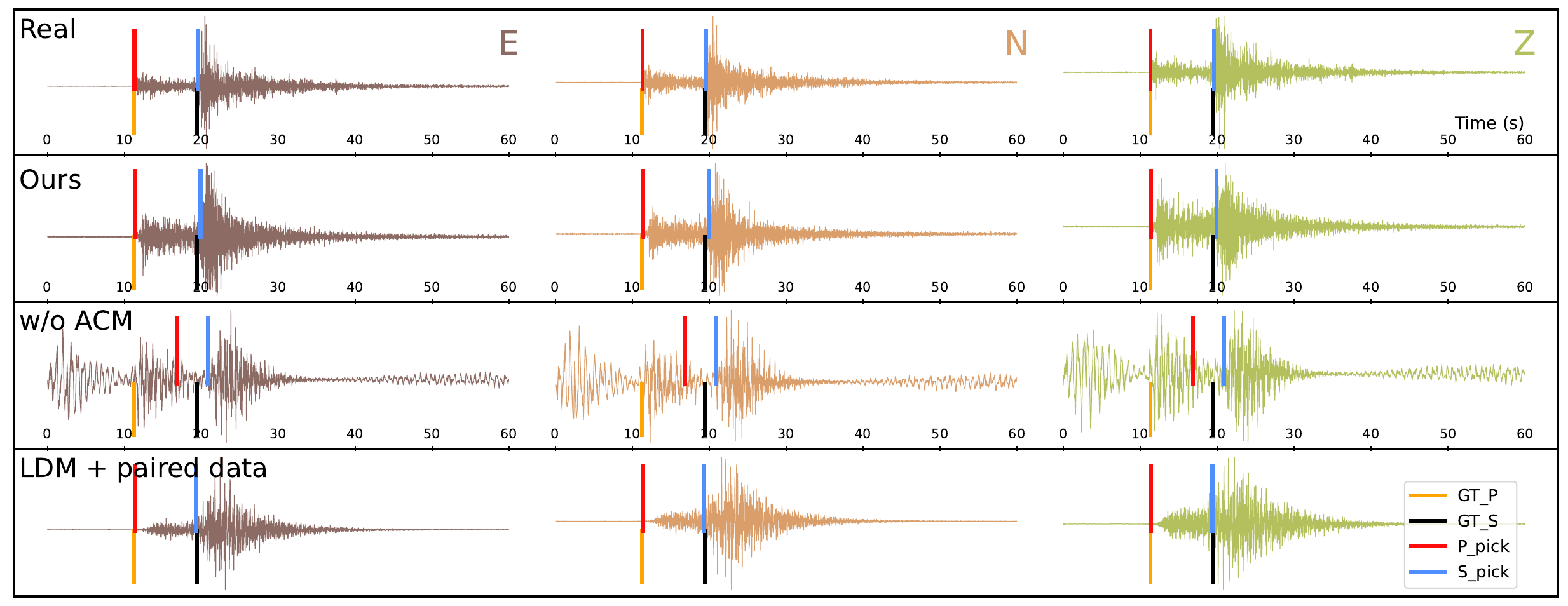}
    \caption{Negative synthesis results of our model and ablation models, compared to the real observation.}
    \label{fig:ablation_bad_159}
\end{figure}

\begin{figure}[H]
    \centering
    \includegraphics[width=0.85\textwidth]{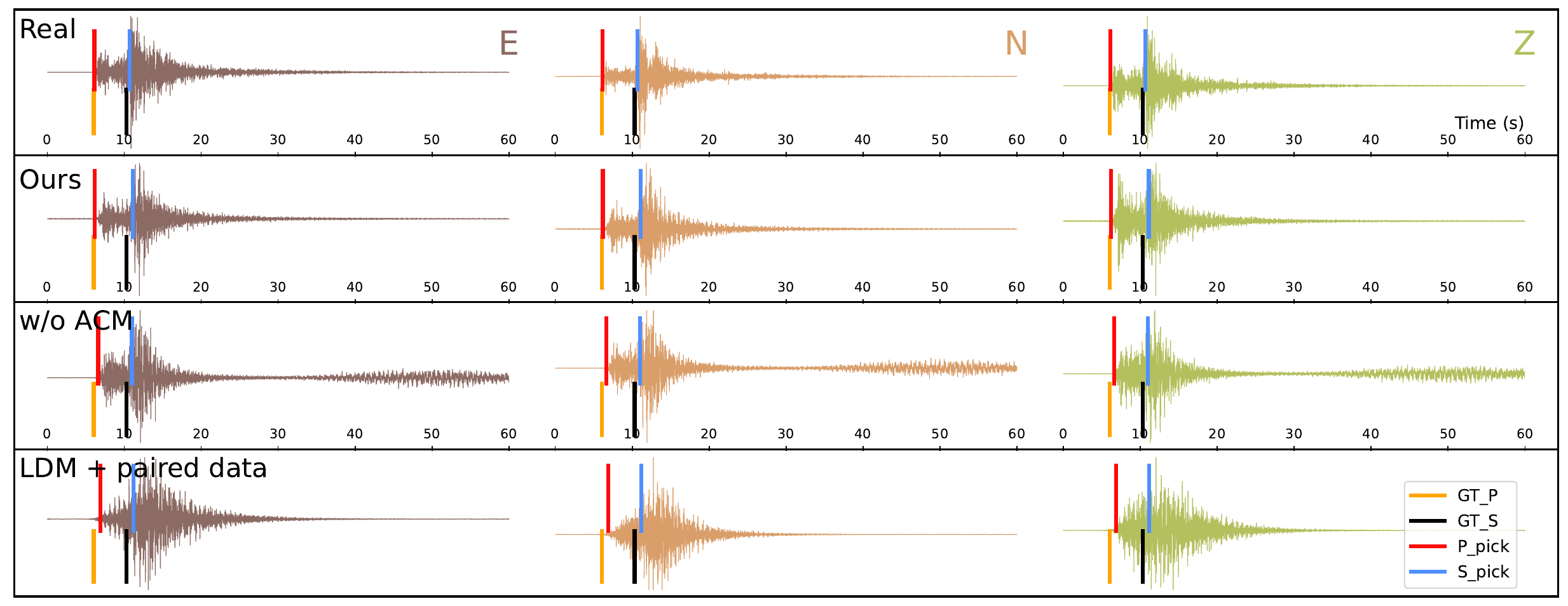}
    \caption{Negative synthesis results of our model and ablation models, compared to the real observation.}
    \label{fig:ablation_bad_254}
\end{figure}

\begin{figure}[H]
    \centering
    \includegraphics[width=0.85\textwidth]{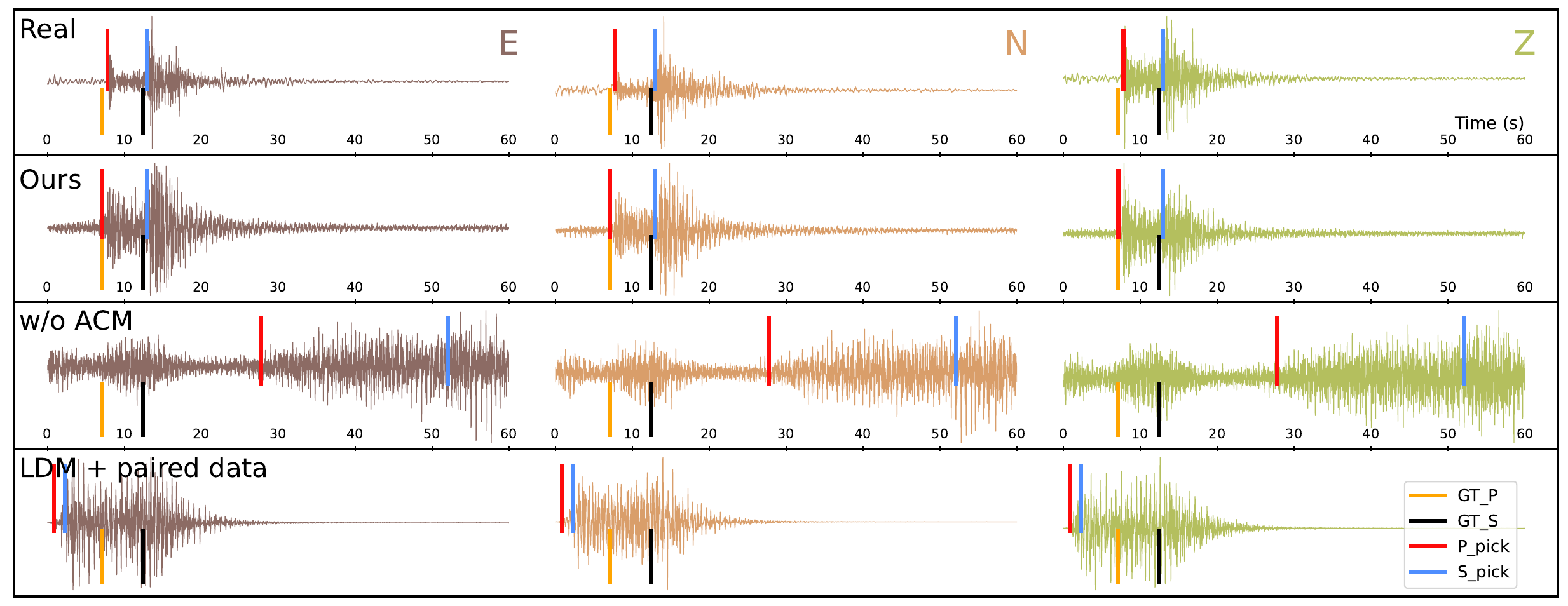}
    \caption{Negative synthesis results of our model and ablation models, compared to the real observation.}
    \label{fig:ablation_bad_258}
\end{figure}

\section{Additional Figures: Waveform and Spectrogram} \label{sec:appendix_waveform_spectrogram}

This section presents the waveforms and spectrograms shown in \cref{fig:waveform} and \cref{fig:spectrograms}. The seismic data we used consist of 3-components, ENZ. Each pair displays the same waveform and spectrogram, with the top representing the real observation and the bottom representing the synthetic generated HEGGS. The red and blue lines on the waveforms indicate the P/S arrival times.

\subsection{SCEDC}

\begin{figure}[H]
    \centering
    \begin{subfigure}{0.7\textwidth}
        \centering
        \includegraphics[width=\textwidth]{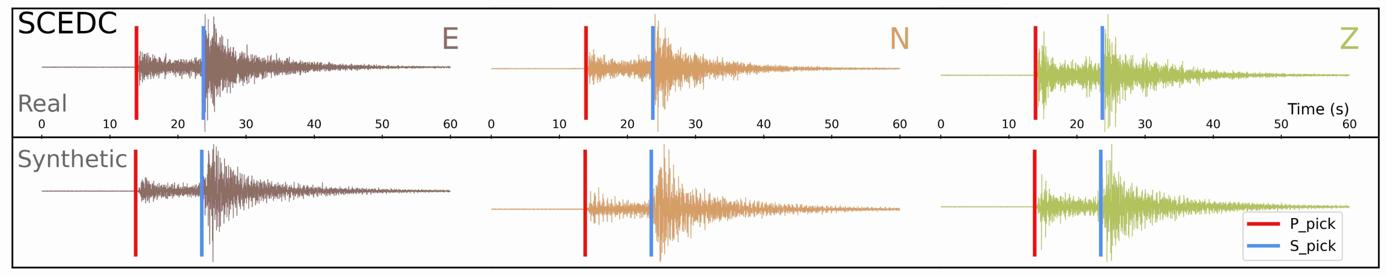}
        \caption{waveform}
    \end{subfigure}
    \begin{subfigure}{0.7\textwidth}
        \centering
        \includegraphics[width=\textwidth]{images/appendix/scedc/scedc_spec_1.pdf}
        \caption{spectrogram}
    \end{subfigure}
        \caption{Synthesis results of our model compared to the real observation.}
        \label{fig:scedc_sample_1}
\end{figure}

\begin{figure}[H]
    \centering
    \begin{subfigure}{0.7\textwidth}
        \centering
        \includegraphics[width=\textwidth]{images/appendix/scedc/scedc10_wave.pdf}
        \caption{waveform}
    \end{subfigure}
    \begin{subfigure}{0.7\textwidth}
        \centering
        \includegraphics[width=\textwidth]{images/appendix/scedc/scedc_spec_10.pdf}
        \caption{spectrogram}
    \end{subfigure}
        \caption{Synthesis results of our model compared to the real observation.}
        \label{fig:scedc_sample_10}
\end{figure}

\subsection{KMA}

\begin{figure}[H]
    \centering
    \begin{subfigure}{0.7\textwidth}
        \centering
        \includegraphics[width=\textwidth]{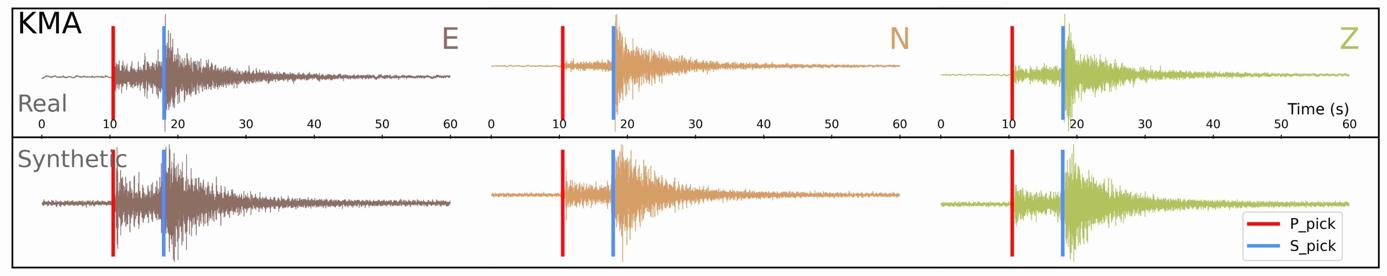}
        \caption{waveform}
    \end{subfigure}
    \begin{subfigure}{0.7\textwidth}
        \centering
        \includegraphics[width=\textwidth]{images/appendix/kma/kma_spec_7.pdf}
        \caption{spectrogram}
    \end{subfigure}
        \caption{Synthesis results of our model compared to the real observation.}
        \label{fig:kma_sample_7}
\end{figure}

\begin{figure}[H]
    \centering
    \begin{subfigure}{0.7\textwidth}
        \centering
        \includegraphics[width=\textwidth]{images/appendix/kma/kma_29_wave.pdf}
        \caption{waveform}
    \end{subfigure}
    \begin{subfigure}{0.7\textwidth}
        \centering
        \includegraphics[width=\textwidth]{images/appendix/kma/kma_spec_29.pdf}
        \caption{spectrogram}
    \end{subfigure}
        \caption{Synthesis results of our model compared to the real observation.}
        \label{fig:kma_sample_29}
\end{figure}

\subsection{INSTANCE}

\begin{figure}[H]
    \centering
    \begin{subfigure}{0.7\textwidth}
        \centering
        \includegraphics[width=\textwidth]{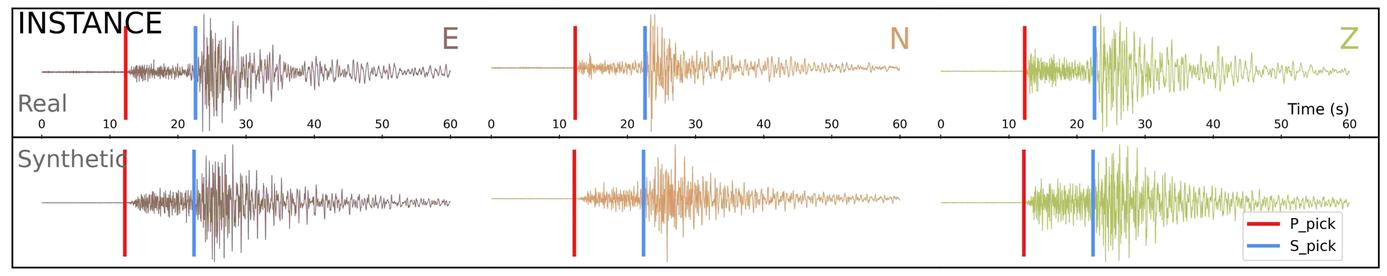}
        \caption{waveform}
    \end{subfigure}
    \begin{subfigure}{0.7\textwidth}
        \centering
        \includegraphics[width=\textwidth]{images/appendix/instance/instance_spec_2.pdf}
        \caption{spectrogram}
    \end{subfigure}
        \caption{Synthesis results of our model compared to the real observation.}
        \label{fig:instance_sample_2}
\end{figure}

\begin{figure}[H]
    \centering
    \begin{subfigure}{0.7\textwidth}
        \centering
        \includegraphics[width=\textwidth]{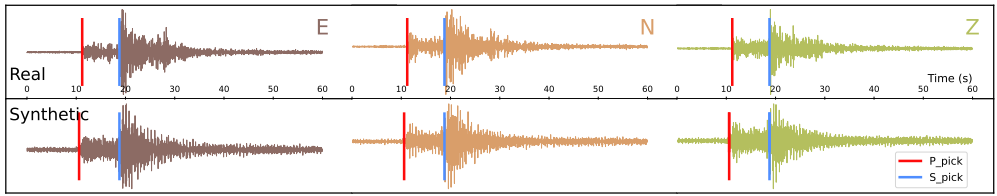}
        \caption{waveform}
    \end{subfigure}
    \begin{subfigure}{0.7\textwidth}
        \centering
        \includegraphics[width=\textwidth]{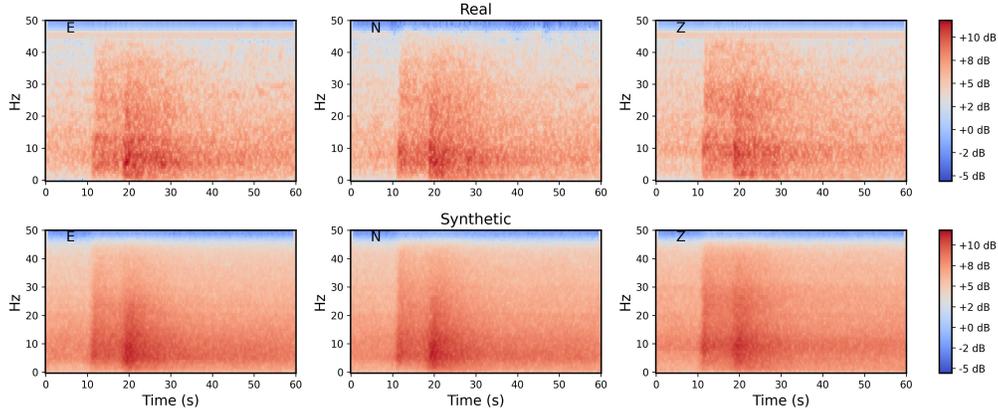}
        \caption{spectrogram}
    \end{subfigure}
        \caption{Synthesis results of our model compared to the real observation.}
        \label{fig:instance_sample_20}
\end{figure}

\section{Additional figures: Section plot} \label{sec:appendix_section_plot}

The section plot is constructed by following process. Initially, a specific earthquake event is chosen, and input data is randomly selected (indicated by the blue line). We set virtual stations established at equidistant intervals from the epicenter, generate waveforms, and plot together with real observations.
The red lines represent ground truth observations and black lines are the synthesized waveforms. Note that the azimuth angle of observations varies, while the synthetic stations are set to have same values. This potentially affect the P/S wave arrivals and lead to mismatch in visualization, but the effect is not considered to be significantly large.

\subsection{SCEDC}

\begin{figure}[H]
    \centering
    \begin{subfigure}[b]{0.3\textwidth}
        \centering
        \includegraphics[width=\textwidth]{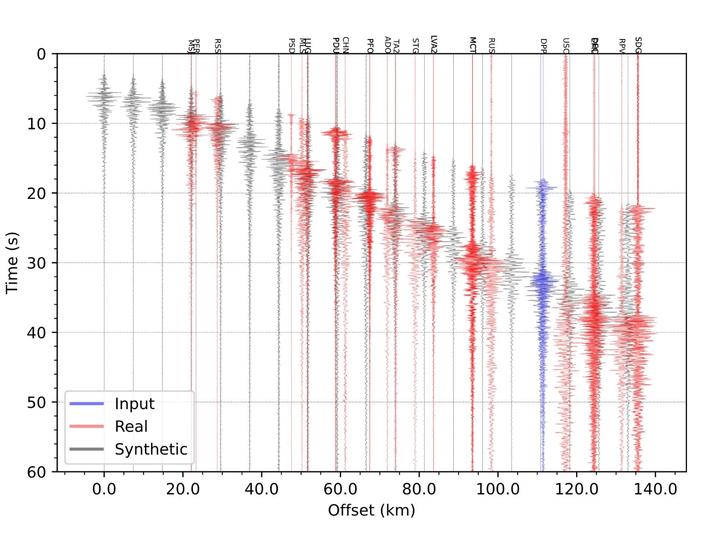}
        \caption{Section plot on E axis}
    \end{subfigure}
    \begin{subfigure}[b]{0.3\textwidth}
        \centering
        \includegraphics[width=\textwidth]{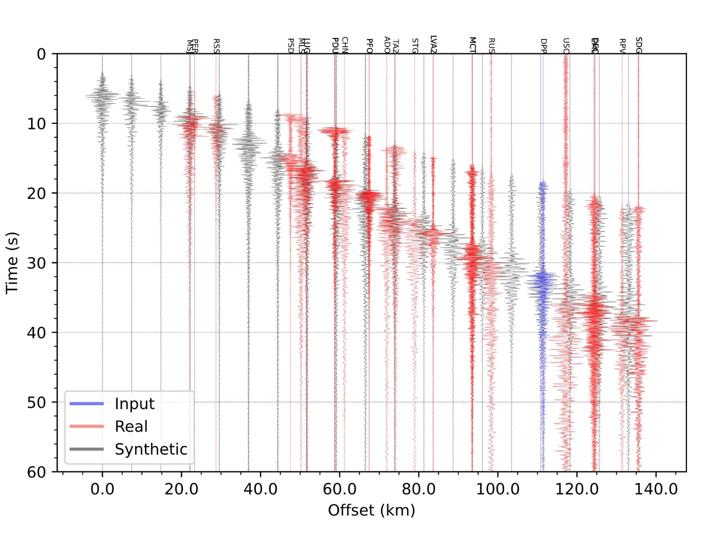}
        \caption{Section plot on N axis}
    \end{subfigure}
    \begin{subfigure}[b]{0.3\textwidth}
        \centering
        \includegraphics[width=\textwidth]{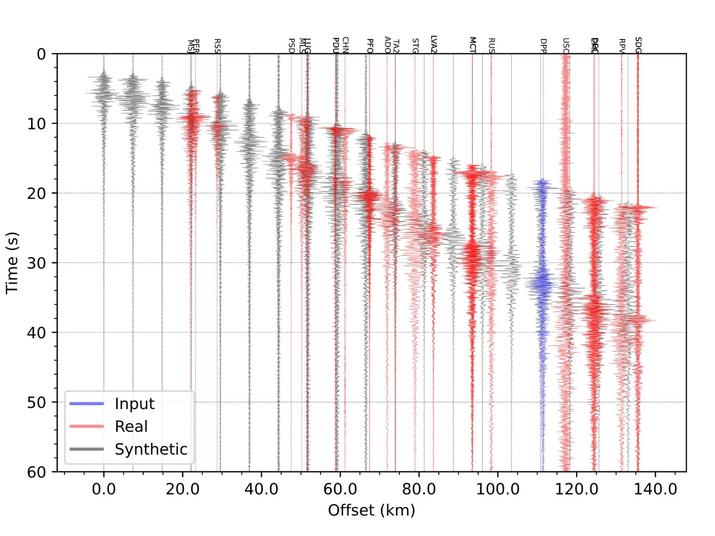}
        \caption{Section plot on Z axis}
    \end{subfigure}
    \caption{Section plot on synthetic stations.}
    \label{fig:supp_section_37374890}
\end{figure}

\begin{figure}[H]
    \centering
    \begin{subfigure}[b]{0.3\textwidth}
        \centering
        \includegraphics[width=\textwidth]{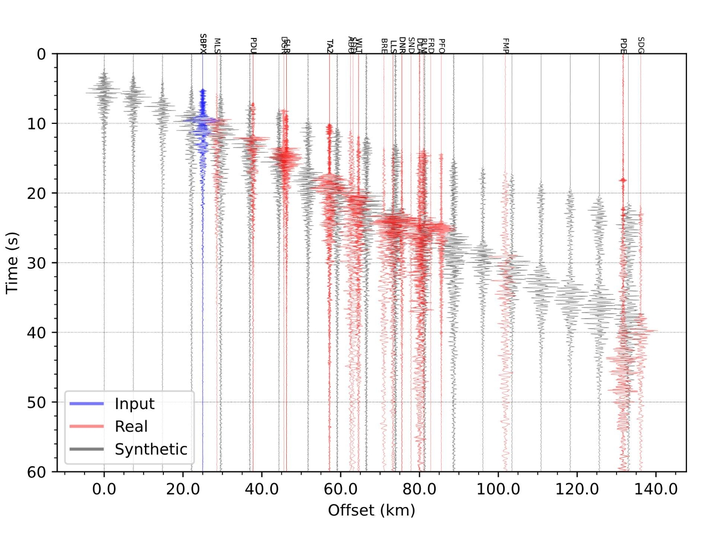}
        \caption{Section plot on E axis}
    \end{subfigure}
    \begin{subfigure}[b]{0.3\textwidth}
        \centering
        \includegraphics[width=\textwidth]{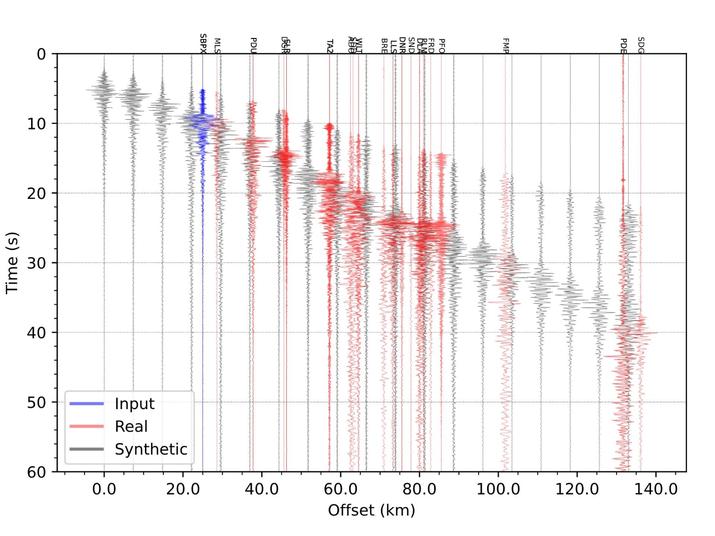}
        \caption{Section plot on N axis}
    \end{subfigure}
    \begin{subfigure}[b]{0.3\textwidth}
        \centering
        \includegraphics[width=\textwidth]{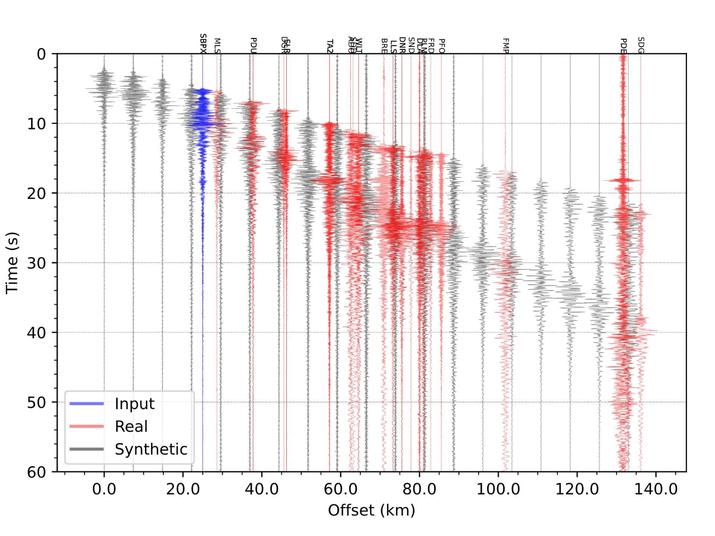}
        \caption{Section plot on Z axis}
    \end{subfigure}
    \caption{Section plot on synthetic stations.}
    \label{fig:supp_section_37713424}
\end{figure}

\subsection{KMA}

\begin{figure}[H]
    \centering
    \begin{subfigure}[b]{0.3\textwidth}
        \centering
        \includegraphics[width=\textwidth]{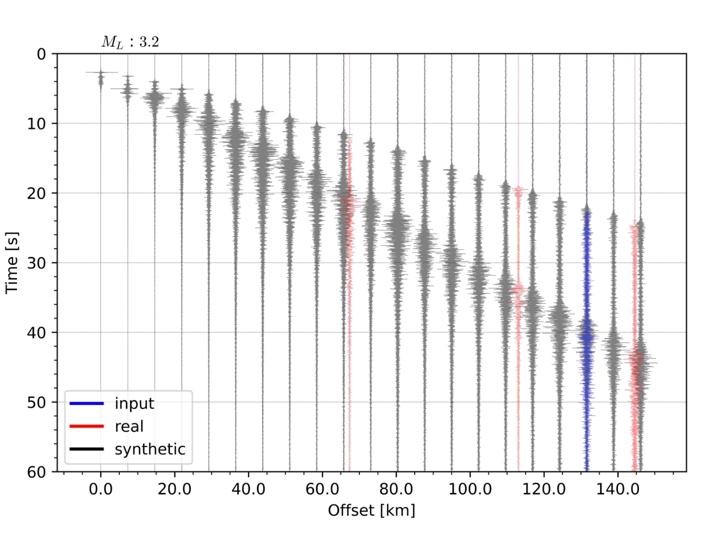}
        \caption{Section plot on E axis}
    \end{subfigure}
    \begin{subfigure}[b]{0.3\textwidth}
        \centering
        \includegraphics[width=\textwidth]{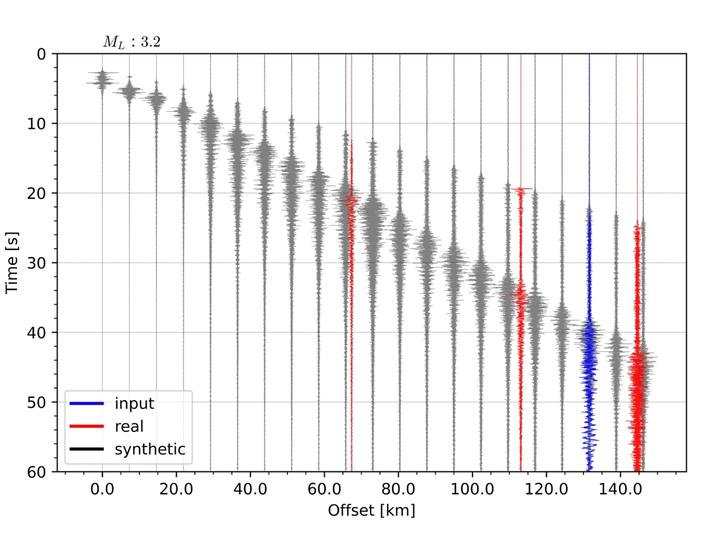}
        \caption{Section plot on N axis}
    \end{subfigure}
    \begin{subfigure}[b]{0.3\textwidth}
        \centering
        \includegraphics[width=\textwidth]{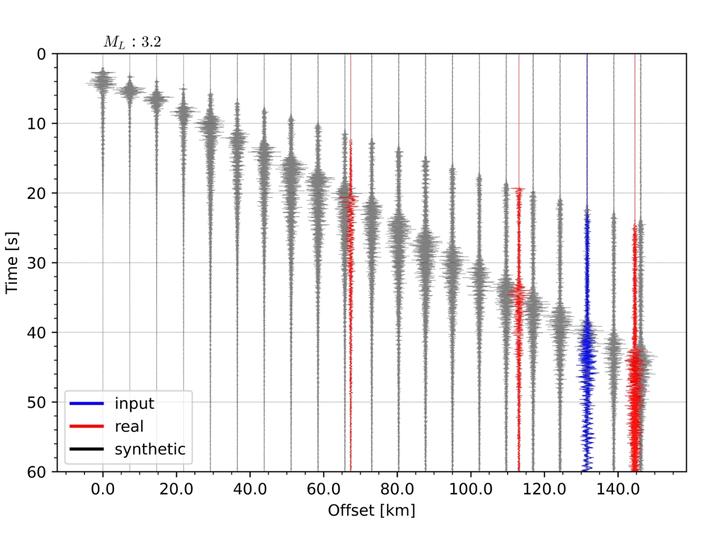}
        \caption{Section plot on Z axis}
    \end{subfigure}
    \caption{Section plot on synthetic stations.}
    \label{fig:supp_section_KS20181128_2145}
\end{figure}

\begin{figure}[H]
    \centering
    \begin{subfigure}[b]{0.3\textwidth}
        \centering
        \includegraphics[width=\textwidth]{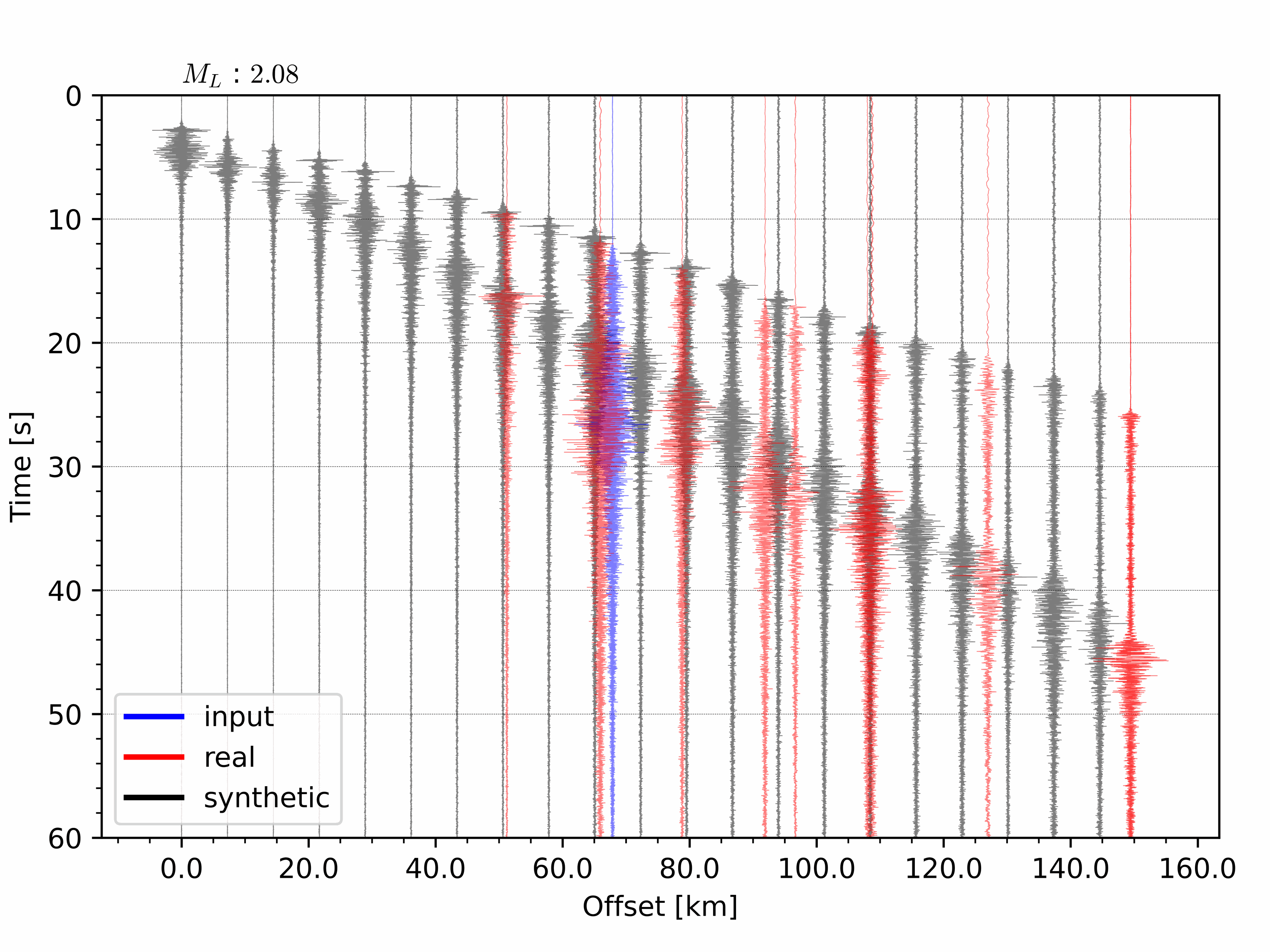}
        \caption{Section plot on E axis}
    \end{subfigure}
    \begin{subfigure}[b]{0.3\textwidth}
        \centering
        \includegraphics[width=\textwidth]{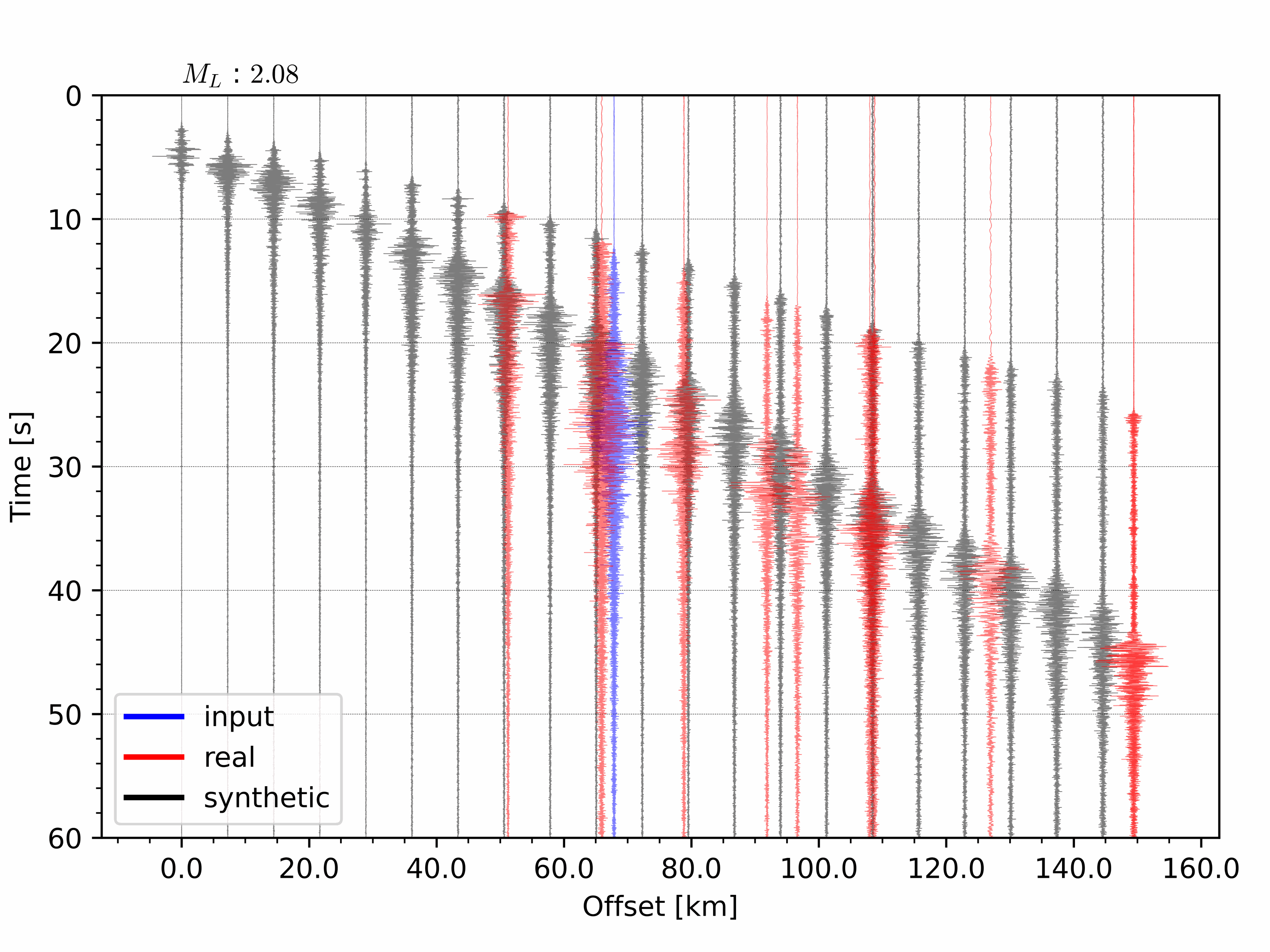}
        \caption{Section plot on N axis}
    \end{subfigure}
    \begin{subfigure}[b]{0.3\textwidth}
        \centering
        \includegraphics[width=\textwidth]{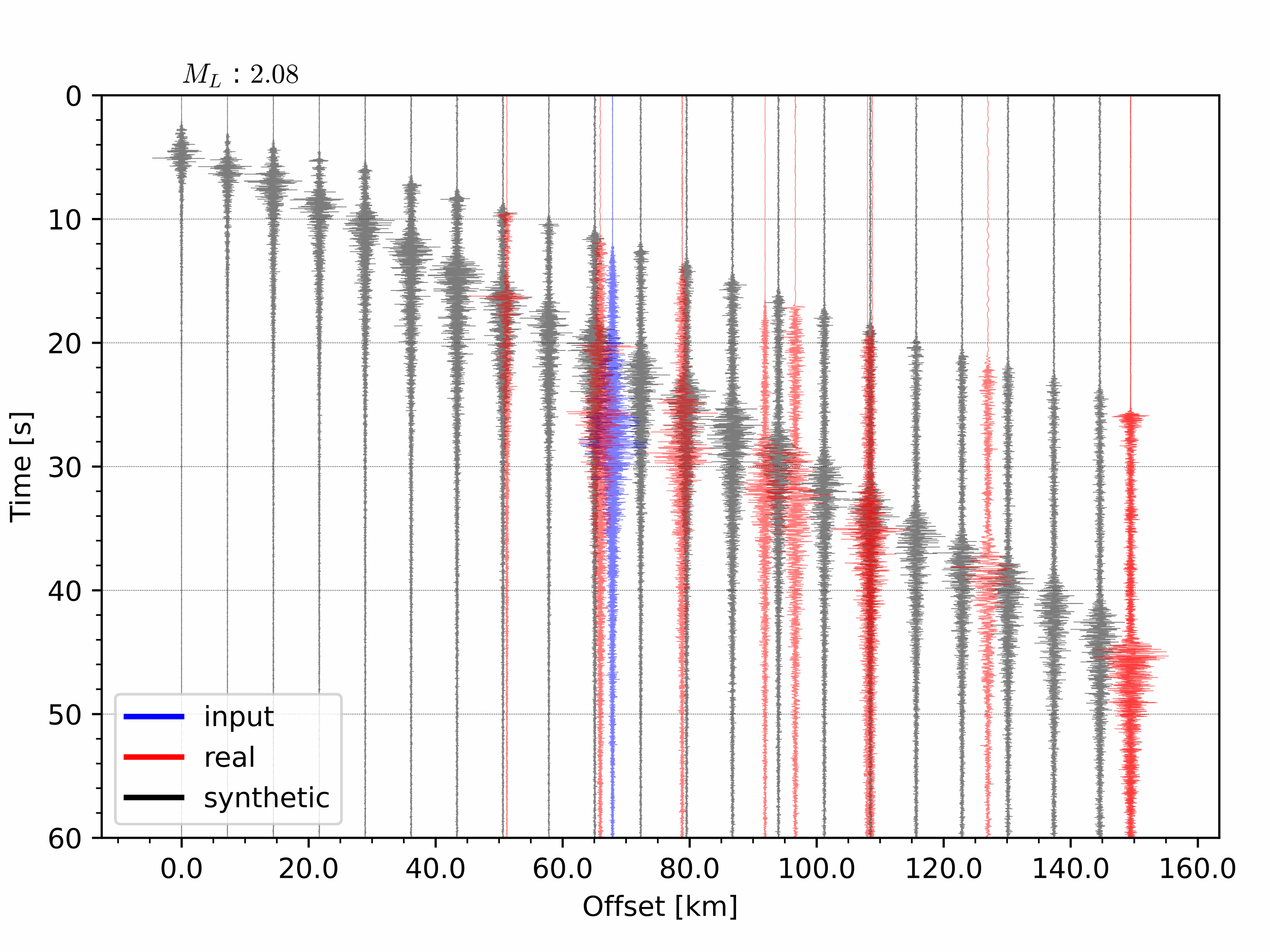}
        \caption{Section plot on Z axis}
    \end{subfigure}
    \caption{Section plot on synthetic stations.}
    \label{fig:supp_section_KS20190721_17371}
\end{figure}

\subsection{INSTANCE}

\begin{figure}[H]
    \centering
    \begin{subfigure}[b]{0.3\textwidth}
        \centering
        \includegraphics[width=\textwidth]{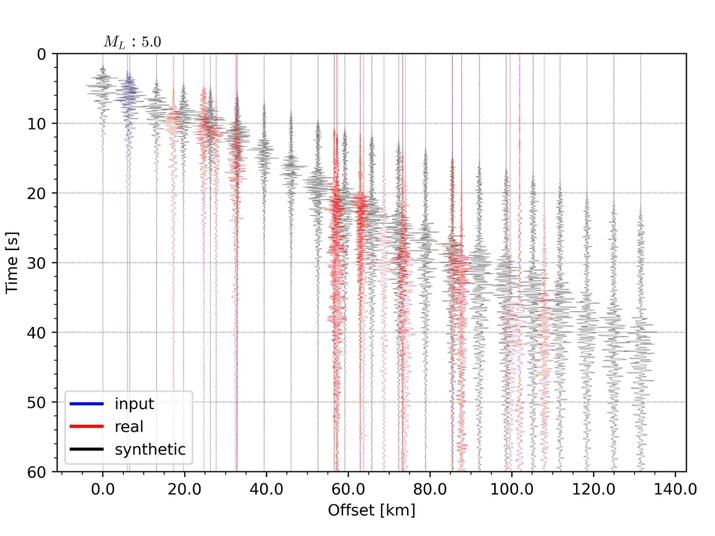}
        \caption{Section plot on E axis}
    \end{subfigure}
    \begin{subfigure}[b]{0.3\textwidth}
        \centering
        \includegraphics[width=\textwidth]{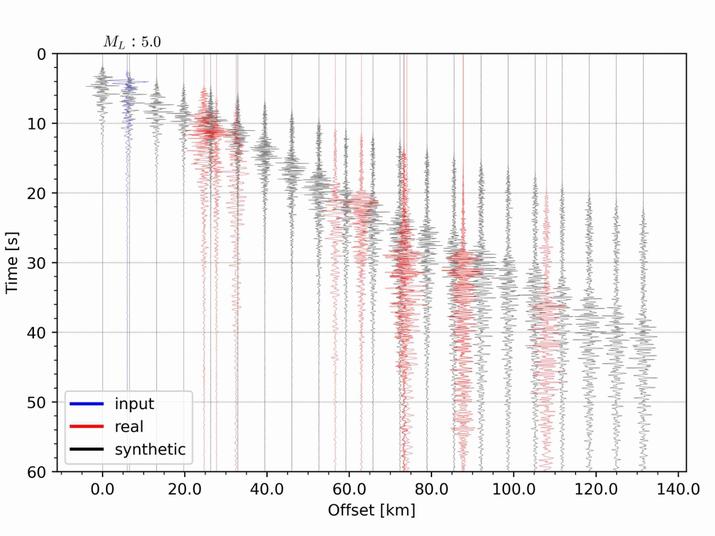}
        \caption{Section plot on N axis}
    \end{subfigure}
    \begin{subfigure}[b]{0.3\textwidth}
        \centering
        \includegraphics[width=\textwidth]{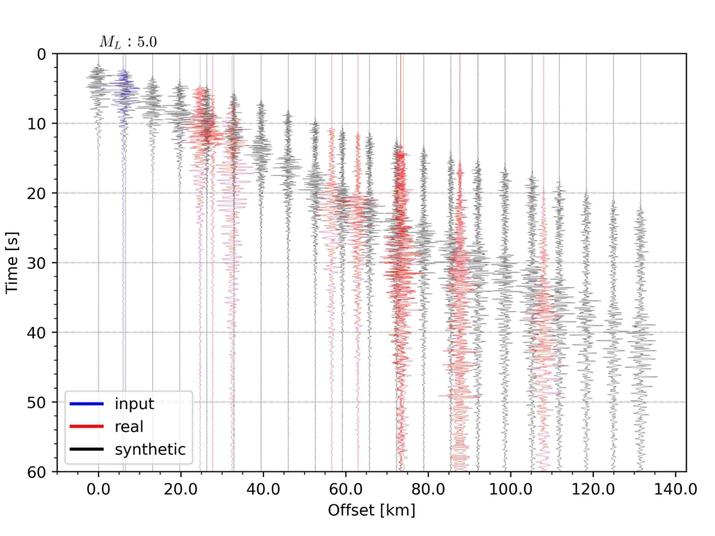}
        \caption{Section plot on Z axis}
    \end{subfigure}
    \caption{Section plot on synthetic stations.}
    \label{fig:supp_section_1921649}
\end{figure}

\begin{figure}[H]
    \centering
    \begin{subfigure}[b]{0.3\textwidth}
        \centering
        \includegraphics[width=\textwidth]{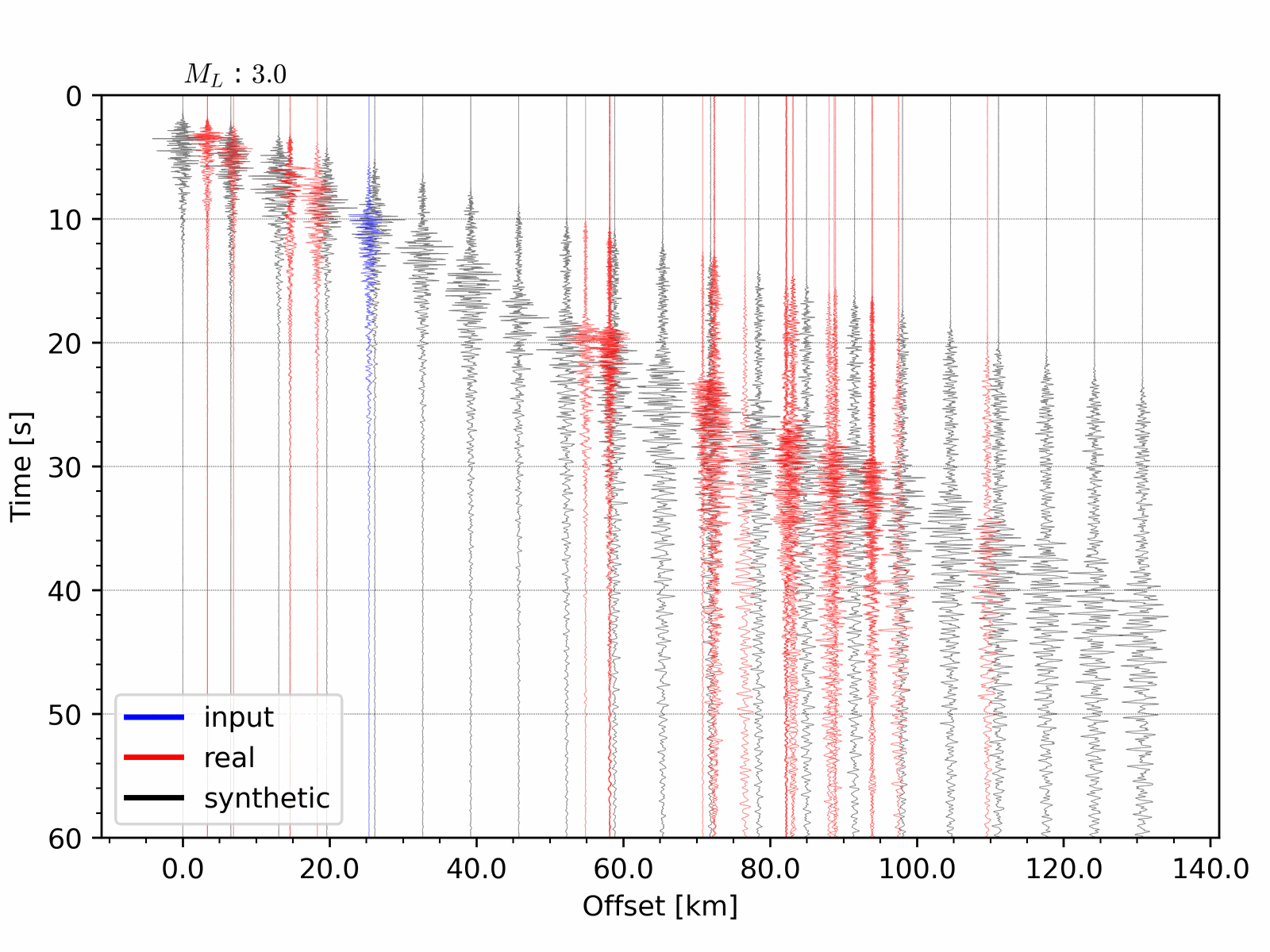}
        \caption{Section plot on E axis}
    \end{subfigure}
    \begin{subfigure}[b]{0.3\textwidth}
        \centering
        \includegraphics[width=\textwidth]{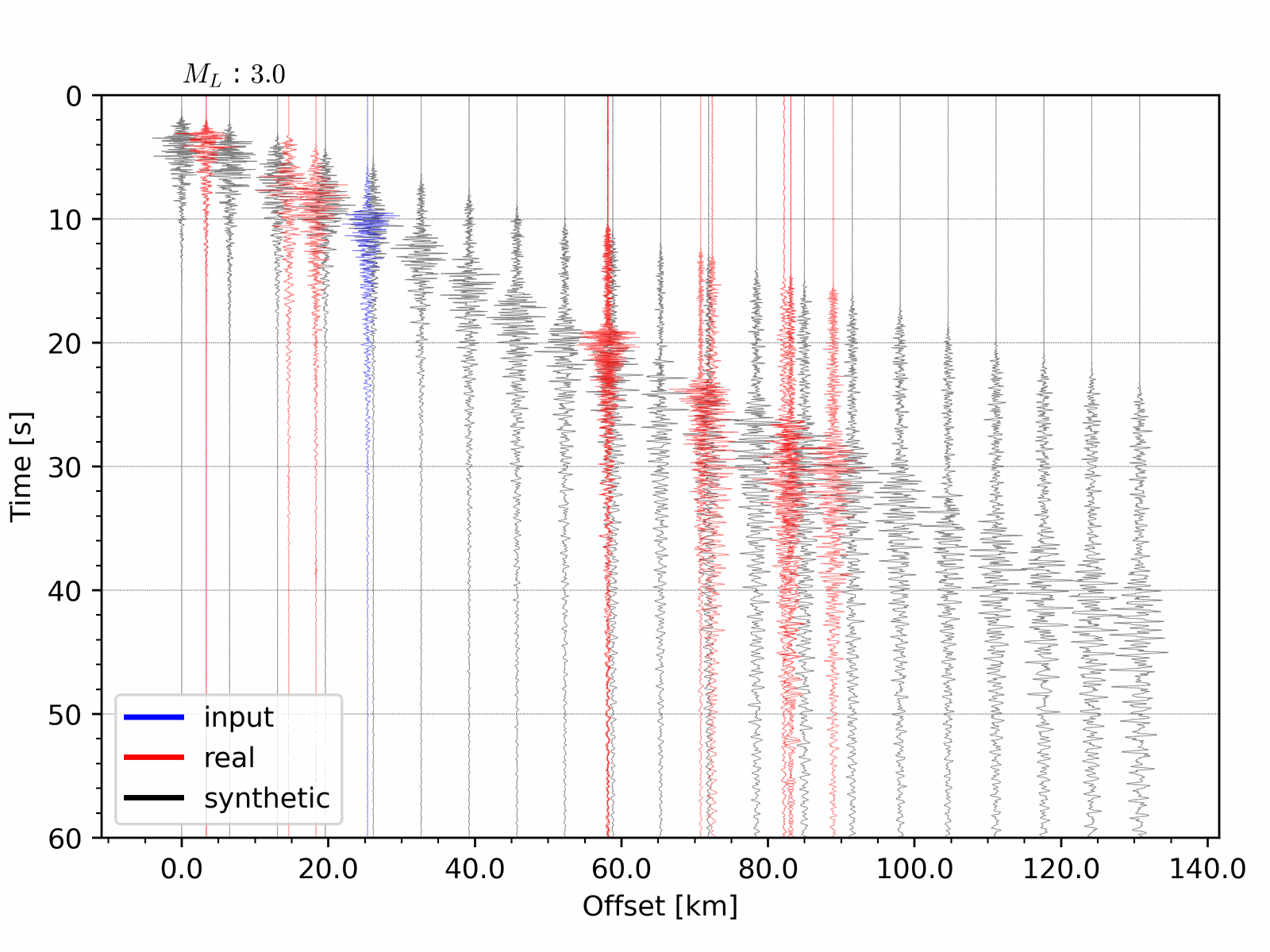}
        \caption{Section plot on N axis}
    \end{subfigure}
    \begin{subfigure}[b]{0.3\textwidth}
        \centering
        \includegraphics[width=\textwidth]{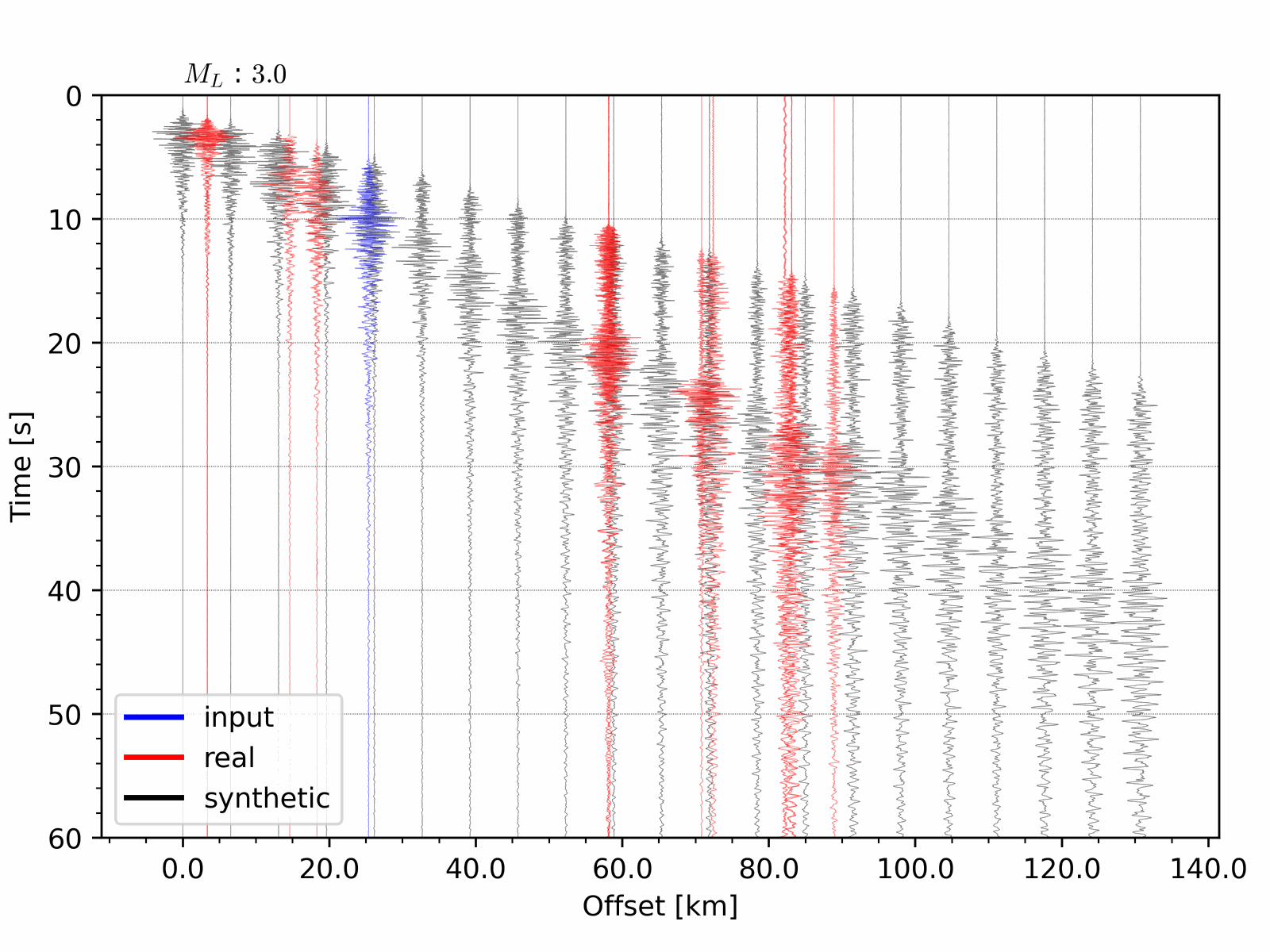}
        \caption{Section plot on Z axis}
    \end{subfigure}
    \caption{Section plot on synthetic stations.}
    \label{fig:supp_section_10000641}
\end{figure}

\newpage
\section{On pair-Exploiting Diffusion Model}\label{appendix:pair-exploiting}
In this appendix, we provide more detailed explanations about the training and inference of HEGGS for the clarification.

\subsection{Remark:Prediction targets of diffusion model}
Referring equation (2),(4),(6) and (7) of \cite{DDPM}, the forward process $q(x_{1:T};X)$ would be given by

\begin{equation}
q(x_t|x_{t-1}) = \mathcal{N}(\sqrt{1-\beta_t}x_{t-1},\beta_tI), q(x_t|x_0) = \mathcal{N}(\sqrt{\overline{\alpha}_{t}}x_0,(1-\overline{\alpha}_t)I).
\end{equation}

and thus we have 
\begin{equation}
    x_t = \sqrt{\overline{\alpha}_{t}}x_0+\sqrt{1-\overline{\alpha}_t}\epsilon \mbox{ where } \epsilon\sim\mathcal{N}(0,1).
\end{equation}

The backward process $q(x_{t-1}|x_{t},x_0)$ would be 
\begin{equation}
    q(x_{t-1}|x_{t},x_0)\sim \mathcal{N}(\tilde{\mu}(x_t,x_0),\tilde{\beta}_tI)
\end{equation}
where
\begin{equation}\label{eqn:diffusion_backward}
    \tilde{\mu}_t(x_t,x_0)=\frac{\sqrt{\overline{\alpha}_{t-1}}\beta_t}{1-\overline{\alpha_t}}x_0+\frac{\sqrt{\alpha_t}(1-\overline{\alpha}_{t-1})}{1-\overline{\alpha_t}}x_t \mbox{  and } \tilde{\beta}_t =\frac{1-\overline{\alpha}_{t-1}}{1-\overline{\alpha}_t}\beta_t.
\end{equation}

In implementation, it is required to find $\tilde{\mu}_t(x_t,x_0)$ term in \cref{eqn:diffusion_backward}.
There are several methods for the prediction, with replacing $x_0$ by estimate $x_\theta(x_t,t)$. The HEGGS directly predicts $x$ in sample space, as it would be more natural since we want to learn the morphology between paired data, compared to the alternatives which predicts the noise $\epsilon$ \cite{DDPM} or $\mathbf{v}$-prediction\cite{vpred}.

\begin{minipage}{0.49\textwidth}
\begin{algorithm}[H]
   \caption{HEGGS training}
   \label{alg:training}
\begin{algorithmic}
   \STATE {\bfseries Input:} Seismic dataset $\mathbb{D}$, diffusion steps $T$
   \REPEAT
   \STATE $(W^{src},W^{tgt},\vec{c}_{tgt})\sim\mathbb{D}$
   \STATE convert $(W^{src},W^{tgt})$ to $(X^{src},X^{tgt})$ 
   \STATE $t\sim Uniform({1,\cdots,T})$
   \STATE $\epsilon\sim\mathcal{N}(0,1)$
   \STATE Take gradient descent step on
   
   \STATE \hskip1.0em $\nabla\|X^{tgt}-\mathcal{D}_{AE}(\mathbf{m}_\theta(z_t^{src},\vec{c}_{tgt},t))\|^2$
   \STATE \hskip1.0em where $z_t^{src}=\sqrt{\overline{\alpha}_t}\mathcal{E}_{AE}(X^{src})+\sqrt{1-\overline{\alpha}_t}\epsilon$
   \UNTIL{converged}

\end{algorithmic}
\end{algorithm}
\end{minipage}
\begin{minipage}{0.49\textwidth}
\begin{algorithm}[H]
   \caption{Generation}
   \label{alg:inference}
\begin{algorithmic}
   \STATE {\bfseries Input:} Diffusion steps $T$,condition vector $\vec{c}_{tgt}$, source waveform $W^{src}$ (optional)
   \IF{$W^{src}$ is given} 
   \STATE convert $W^{src}$ to spectrogram $X^{src}$
   \STATE $z_T=\mathcal{E}_{AE}(X^{src})$
   \ELSE
     \STATE sample $z_T\sim\mathcal{N}(0,1)$
    \ENDIF
   \FOR{$t=T,\cdots,1$}
   \STATE sample $\mathbf{z}\sim\mathcal{N}(0,1)$
   \STATE compute $\tilde{z} = \mathbf{m}_\theta(z_t,\vec{c}_{tgt},t)$
   \STATE compute $z_{t-1}=\tilde{\mu}_t(z_{t},\tilde{z})+\sqrt{\tilde{\beta}_t}\mathbf{z}$ (Eq. \ref{eqn:diffusion_backward})
   \ENDFOR
   \STATE $X^{tgt} = \mathcal{D}_{AE}(z_0)$
   \STATE Convert $X^{tgt}$ to waveform $W^{tgt}$
   \STATE {\bfseries Return:}  $W^{tgt}$
\end{algorithmic}
\end{algorithm}
\end{minipage}

\subsection{Training with pairs}

As described in \cref{sec: method}, we consider the paired data $(X^{src},X^{tgt})$ with corresponding condition vector $\vec{c}_{src}$ and $\vec{c}_{tgt}$. Note that $\vec{c}_{src}$ is not in use. 

Since $X^{src}$ and $X^{tgt}$ are the observations of same earthquake, we make assumption that there exist a morphology $\eta$ which maps the latent $x_t^{src}$ of $X^{src}$ at time $t$, to $x_t^{tgt}$ using $\vec{c}_{tgt}$, as a random variable. We formulate this assumption with \cref{eqn:src-to-tgt-eta}, as follows:
\begin{equation}
    \eta(x_t^{src},\vec{c}_{tgt},t)\sim q(x_t^{tgt}|X^{tgt}) \tag{\ref{eqn:src-to-tgt-eta}}
\end{equation}

This assumption includes the intuition that the broadband waveform signal is a combination of earthquake information, which is considered to be included in $X^{src}$, and local geological features near observatory, encoded by positional information from $\vec{c}_{tgt}$.

For training, we aim to train the neural network $\mathbf{m}_\theta$ which is a composition of $\eta$ and denoising model $\mathbf{x}_\theta$. Precisely, $\mathbf{m}_\theta$ would be written by 
\begin{equation}
    \mathbf{m}_\theta(x,\vec{c},t) =  \mathbf{x}_\theta(\eta(x,\vec{c},t),\vec{c},t).
\end{equation}

Since $\eta(x_t^{src},\vec{c}_{tgt},t)=x_t^{tgt}$, we have $\mathbf{m}_\theta(x_t^{src},\vec{c}_{tgt},t)=\mathbf{x}_\theta(x_t^{tgt},\vec{c}_{tgt},t)$ for the paired latents $(x_t^{src},x_t^{tgt})$, the loss function of diffusion model \cref{eqn:loss_DM_vanilla} would be equivalent to \cref{eqn:loss_DM_paired}:
\begin{equation}
    \mathcal{L}_{DM}' = \mathbb{E}_{(X^{src},X^{tgt},\vec{c}_{tgt}),\epsilon,t}\|X^{tgt}-\mathbf{m}_\theta(x_t^{src},\vec{c}_{tgt},t)\|^2
\end{equation}

After that, we consider same procedure in latent space (the $z_t^{tgt}$ for the clarification) with autoencoder consist of the encoder $\mathcal{E}_{AE}$ and decoder $\mathcal{D}_{AE}$, we obtain the loss function \cref{eqn:loss-ours}, with end-to-end training. 

\begin{equation}
    \mathcal{L}_{ours}  :=\mathbb{E}_{(X^{src},X^{tgt},\vec{c}_{tgt}),\epsilon,t} \|X^{tgt}-\mathcal{D}_{AE}(\mathbf{m}_\theta(z_t^{src},\vec{c}_{tgt},t))\|^2 \tag{\ref{eqn:loss-ours}}
\end{equation}

In \cref{alg:training}, we present an training algorithm for the HEGGS training with $\mathcal{L}_{ours}$. The paired waveforms and corresponding condition vector of target waveform would be sampled from the dataset, and the gradient descent would update all modules $\mathbf{m}_\theta,\mathcal{E}_{AE}$ and $\mathcal{D}_{AE}$ together. 

\begin{remark}
For the training process of diffusion model with \cref{eqn:loss-ours}, several details below are considered for the loss and model design.
\begin{enumerate}
    \item During the training, the noise is designed to be added to the $Z^{src}$ instead of $Z^{tgt}$. This would provide robustness against site-specific noise which already included in observation $W^{src}$ and its latent vector $Z^{src}$.
    \item When $t$ is small, $z_t^{src}$ would be almost same to $Z^{src}$ (this is also because $X^{src}$ itself is already noisy) and thus the model would learn the transformation $\eta$ with more attention.
    \item Regarding the intuition that $z_t^{src}$ and $z_t^{tgt}$ will be identified (in distribution) when $t$ is sufficiently large, the training loss \cref{eqn:loss-ours} would be equivalent to the conventional training loss for $\mathbf{x}_\theta$ training when we disregard the end-to-end training. Hence, the model learns to generate from the noise w/o $W^{src}$ too, during the training.
    \item Since $\eta$ and $\mathbf{m}_\theta$ does not take $\vec{c}_{src}$ as input. 
    Therefore the model learns to extract common information from $z_t^{src}$ through multiple pairs of observations of same earthquake during training, regardless the local information (encoded by location) of observatory.
    This makes the model can handle $z_t^{tgt}$ as a input too, since it shares the information of earthquake. 
\end{enumerate}
\end{remark}

\subsection{Inference w/o $W^{src}$}

Although the diffusion model is trained with paired data and takes $W^{src}$ as an input, our model is capable to synthesize seismic waveform without the observation $W^{src}$.

Since $\eta$ is defined to map the source latent $z_t^{src}$ to target latent $z_t^{tgt}$, it also maps the target latent to itself, in distribution. Precisely, we can write
\begin{equation}
    \eta(z_t^{tgt},\vec{c}_{tgt},t) = z_t^{tgt}
\end{equation}
and thus the output of neural network would be
\begin{equation}
    \mathbf{m}_\theta(z_t^{tgt},\vec{c}_{tgt},t) = \mathbf{x}_\theta(\eta(z_t^{tgt},\vec{c}_{tgt},t),\vec{c}_{tgt},t) = \mathbf{x}_\theta(z_t^{tgt},\vec{c}_{tgt},t)
\end{equation}

Therefore, we can use conventional reverse process
\begin{equation}
    z_{t-1}^{tgt} = \tilde{\mu}_t(z_t^{tgt},\textbf{m}_\theta(z_t^{tgt},\vec{c}_{tgt},t))+\sigma_t\textbf{z}, \textbf{z}\sim N(0,I)
\end{equation}
even if $z_T^{tgt}$ is the gaussian noise sampled from $\mathcal{N}(0,1)$.

In \cref{alg:inference}, we summarize the generation process of our model. Note that the diffusion steps are equivalent to LDM\cite{LDM} when $W^{src}$ is not given.

\end{document}